\documentclass[11pt,draftcls,onecolumn]{IEEEtran}
\usepackage{cite}

\ifCLASSINFOpdf
\usepackage[pdftex]{graphicx}
\else
\usepackage[dvips]{graphicx}
\fi

\usepackage{amssymb}
\usepackage{amsmath}
\usepackage{color, graphicx}
\usepackage{array}
\usepackage{algorithm}
\usepackage{algorithmic}
\usepackage{url}

\begin{document}
\title{Task-Driven Adaptive Statistical Compressive Sensing of Gaussian Mixture Models}

\author{Julio~M.~Duarte-Carvajalino,
        Guoshen~Yu,
        Lawrence~Carin, ~\IEEEmembership{Fellow~Member,~IEEE},
        and~Guillermo~Sapiro,~\IEEEmembership{Senior~Member,~IEEE}
\thanks{J. M. Duarte-Carvajalino, G. Yu, and G. Sapiro are with the Department
of Electrical and Computer Engineering, University of Minnesota, Minneapolis,
MN, 55455-0436 USA (e-mail: guille@umn.edu).}
\thanks{L. Carin is with the the Department
of Electrical and Computer Engineering, Duke University, Durham,
NC, 27708-0291 USA.}}

\markboth{IEEE Transactions on Signal Processing}{Duarte \MakeLowercase{\textit{et al.}}}
\maketitle

\begin{abstract}
A framework for adaptive and non-adaptive statistical compressive sensing is developed, where a statistical model replaces the standard sparsity model of classical compressive sensing. We propose within this framework optimal task-specific sensing protocols specifically and jointly designed for classification and reconstruction. 
A two-step adaptive sensing paradigm is developed, where online sensing is applied to detect the signal class in the first step, 
followed by a reconstruction step adapted to the detected class and the observed samples.
The approach is based on information theory, here tailored for Gaussian mixture models (GMMs), where an information-theoretic objective relationship between the sensed signals and a representation of the specific task of interest is maximized. 
Experimental results using synthetic signals, Landsat satellite attributes, and natural images of different sizes and with different noise levels show the improvements achieved using the proposed framework when compared to more standard sensing protocols. The underlying formulation can be applied beyond GMMs, at the price of higher mathematical and computational complexity. 
\end{abstract}


\IEEEpeerreviewmaketitle

\section{Introduction}
\label{Introduction}
\IEEEPARstart{C}{ompressive} sensing (CS) theory states that signals $\mathbf{x} \in \mathbb{R}^N$ that have a sparse or compressible  representation on a dictionary $\mathbf{D}$ can be sensed using far less linear measurements, $\mathbf{y} = \mathbf{\Phi x} \in \mathbb{R}^M, M \ll N$, than those required by the Shannon-Nyquist theorem, with minimum loss of information \cite{CandesWakin}; it is assumed $\mathbf{x} = \mathbf{D} \boldsymbol{\alpha}$, where $\boldsymbol{\alpha}$ is sparse or nearly sparse. In addition to the sparsity of the signals, CS requires the sensing matrix $\mathbf{\Phi}$ be as incoherent (uncorrelated) as possible with the dictionary \cite{CandesWakin}. A key property in CS is the Restrictive Isometry Property (RIP) that enforces incoherence and ensures robustness of the reconstruction \cite{CandesWakin, Peyre}.  In particular, its has been shown that random sensing matrices satisfy the RIP with overwhelming probability \cite{CandesWakin, Candes2005, Candes2006, CandesSpain}. For specific types of signals (represented by their associated, often learned, dictionary), deterministic sensing matrices can be designed \cite{DeVore, Calderbank},  or better yet learned, leading to significant improvements over random sensing matrices \cite{Elad, Peyre, Duarte2009, Yonina, Duarte2012}. 

Off-the-shelf dictionaries are not flexible enough to represent the large variability found in natural signals \cite{Julien}. Learned overcomplete dictionaries can capture better this variability, but the optimization over general unstructured overcomplete dictionaries is often expensive and unstable due to the fact that the search space increases combinatorially with the number of atoms (columns) in the dictionary \cite{Guoshen2010, MarcoDuarte}. Structured overcomplete (learned) dictionaries have been proposed to reduce the size of the search space, improving the sparse representation of complex signals \cite{Lu, Blumensath, Guoshen2010, MarcoDuarte}.       

Reconstruction of CS signals is usually performed via nonlinear optimization strategies such as regularized orthogonal matching pursuit (OMP) \cite{Deanna} and $\ell_1$ convex optimization \cite{CandesWakin, Candes2005, Candes2006, CandesSpain, Donoho}. A piecewise linear inversion model (PLM) was recently introduced \cite{Guoshen2010, Guoshen2011} (see also \cite{Chen}), based on the maximum \emph{a posteriori} expectation-maximization method (MAP-EM) for signals following a learned statistical Gaussian Mixture Model (GMM), which is a case of structured sparsity. The PLM has been shown to be effective and computationally efficient to reconstruct signals degraded by noise, blurring, sub-sampling, or any other linear filters such as CS random matrices. Theoretical analysis \cite{Guoshen2011, Chen} (which mostly considers random sensing matrices) indicates numerous advantages of such a statistical model compared to standard deterministic sparsity models.

The original CS framework advocates the use of non-adaptive linear measurements. Adaptive CS \cite{ACS, Haupt2009, Haupt2011, Aldroubi} has been recently introduced, where each new measurement uses the information obtained from the previous measurements, focusing on subspaces that are more likely to contain true signal components \cite{Haupt2009, Haupt2011}. A generalization of the adaptive CS theory is the adaptive task specific imaging (ATSI) framework, \cite{Neifeld2008}, where the task can be reconstruction \cite{Neifeld2010, Carin2012}, classification \cite{Neifeld2009,Carin_private}, or target detection \cite{Neifeld2008}. ATSI adaptively selects the new measurements that maximize the mutual information between the CS signals and a representation of the specific task of interest (labels for instance in classification), thereby connecting information theory with compressed sensing. 

In \cite{Duarte2012}, we extended \cite{Duarte2009} and proposed a non-adaptive (batch) sensing matrix for statistical CS (SCS) of Gaussian mixture models (GMMs), shown to outperform random sensing matrices. However, the non-adaptive sensing matrix proposed in \cite{Duarte2012} did not exploit the structure of the dictionary, nor the model employed. In this work, we first propose a new optimal sensing matrix specifically designed for SCS of GMMs. This sensing matrix can be used for classical non-adaptive CS, or as explained in Section \ref{sec_adaptive}, in the first step of a novel two-step adaptive statistical CS (ASCS) here introduced and described next. 

Inspired in part by ATSI \cite{Neifeld2008, Neifeld2009}, we propose an adaptive two-step SCS framework, tailored to the learned GMM. In the first step, the task is classification/detection, where we can either use non-adaptive measurements (computationally cheaper, see Sections \ref{sec_nonAdaptive} and \ref{sec_adaptive}) or we can online add adaptive measurements (computationally more expensive, see Section \ref{sec_adaptive}) that maximize an information-theoretic objective function between the CS signals and the classes (Gaussians), while considering the measurements made so far (the measurements can be optimized in batch mode or one at a time). Once the Gaussian model has been estimated in the first step, the measurements on the second step are chosen (in one offline optimally computed block) either from a non-adaptive block optimal for the proper Gaussian model, or from an adaptive block that maximizes the mutual information between the CS signals and the original signal (for reconstruction purposes), taking into account both the previous measurements and the detected class. Hence, several degrees of adaptivity are possible (Section \ref{sec_adaptive}), offering a great deal of flexibility. The computational complexity of the different options is presented as well. As we will later see, this two-step adaptive CS paradigm improves reconstruction accuracy, compared to using non-adaptive measurements. 
In addition, we use sequential hypothesis testing \cite{SHT1, SHT2, Neifeld2009} to automatically determine when to stop acquiring new measurements for classification purposes (first step), and automatically switch to the second step, where the focus is on reconstruction, or we stop the acquisition all together if the detected class is not of interest.

Related works, such as ATSI \cite{Neifeld2008, Neifeld2009}, and more recently \cite{Carin2012, Carin_private}, consider a single adaptive step. In particular \cite{Neifeld2008, Neifeld2009} use parametric models, where the variables of interest are considered as fixed unknown parameters, and the only source of variability is the additive random noise. Here, the signals are realizations of multidimensional Gaussian random variables, within the GMM. In \cite{Carin2012}, the authors present batch and adaptive CS matrices, based on information theory and its relationship with the minimum mean squared error (MMSE) \cite{Guo, Verdu}. For the specific case of GMMs, additive white Gaussian noise, and a signal with a known Gaussian distribution, the optimal sensing matrix can be constructed by taking the first $M$ eigenvectors of the corresponding covariance matrix \cite{MallatBook, Carin2012}. This result is included here (see Section \ref{sec_nonAdaptive}), as a non-adaptive second step, where we already have an estimate of the identity of the Gaussian (we do not know the identity of the Gaussian in the first step). In \cite{Carin2012, Carin_private}, the authors use general (Gaussian or not) signal models to obtain general sensing matrices, although most of the proposed solutions require expensive Monte Carlo simulations in order to compute expectations. Alternatively, the authors also propose to use a different measure of information, the Renyi entropy \cite{MI_book}, elegantly leading to a closed-form solution for the gradient of the Renyi entropy and the GMM. Here, we exploit a different well-known measure of information ($\mu$-measure, see Section\ref{sec_adaptive}), and provide also a closed-form solution for its gradient in the first step (class detection) and a non-iterative closed-form solution for the second step (reconstruction). The main differences of this work with the seminal works \cite{Neifeld2008, Neifeld2009, Carin2012, Carin_private} are the general two-step framework, the specific SCS models employed, and the mathematical simplicity of the derived equations resulting from the proposed measures. 

In summary, the main contribution of this work is to provide a variety of statistical CS options ranging from a simple non-adaptive block SCS framework for GMMs, to several CS configurations offering different degrees of adaptivity within the proposed two-step (model detection and then reconstruction) SCS framework. Most of these configurations can be done either offline or online. Despite the fact that we use here a GMM, the proposed approach can be extended to non-Gaussian models at the cost of higher mathematical and computational complexity.  

Section \ref{sec_SCS} briefly reviews the statistical compressive sensing (SCS) of GMMs framework and the associated PLM reconstruction. Section  \ref{sec_nonAdaptive} introduces the optimal non-adaptive statistical CS approach  for GMMs. Section \ref{sec_adaptive} presents the several adaptive options within the two-step adaptive statistical CS paradigm, focusing on the most adaptive cases. Section \ref{sec_experiments} presents the highlights of the experiments with synthetic and real signals (patches from natural images and Landsat satellite attributes), where the different degrees of adaptivity are compared. Conclusions of this work are presented on Section \ref{sec_conclusion}, and the Appendix contains all the mathematical derivations that were not included in the text for clarity of the exposition. Numerous additional experimental results are included in the supplementary material.   
 
\section{Statistical Compressive Sensing of Gaussian Mixture Models}
\label{sec_SCS}
Let us assume that there exist $G$ Gaussian distributions such that the signals of interest $\mathbf{x} \in \mathbb{R}^N$ can be modeled as a mixture of Gaussians
\begin{align}
p(\mathbf{x}) &= \sum^G_{g=1} p(g) \mathcal{N} (\mathbf{x} | \boldsymbol{\mu}_g, \mathbf{\Sigma}_g), \nonumber \\
\mathcal{N} (\mathbf{x} | \boldsymbol{\mu}_g, \mathbf{\Sigma}_g) &= \frac{1}{ \sqrt { (2 \pi)^n \arrowvert  \mathbf{\Sigma}_g  \arrowvert } } 
			\exp \Big( -\frac{1}{2} (\mathbf{x}-\boldsymbol{\mu}_g)^T \mathbf{\Sigma}^{-1}_g (\mathbf{x}-\boldsymbol{\mu}_g) \Big ), 
			~ g \in \{1, \ldots, G\},
\label{Eq1}
\end{align}
\noindent where $p(g)$ is the probability that $\mathbf{x}$ is a realization of the $g$-th Gaussian distribution, and $\boldsymbol{\mu}_g \in \mathbb{R}^N$, $\mathbf{\Sigma}_g$ correspond respectively to the mean and $N$$\times$$N$ covariance matrix of the $g$-th Gaussian distribution. Even more, let us assume that a given $\mathbf{x}$ is associated with {\it exactly} one (one-block sparsity) of the $G$ mixture components, and mixture component $g$ is selected with probability $p(g)$. While it is straightforward to extend this to a mixture of two or more Gaussian distributions (beyond one-block sparsity), as indicated in \cite{Guoshen2010, Guoshen2011}, increasing the complexity of the model does not necessarily improve the reconstruction of the signals.

The corresponding structured overcomplete dictionary for this model is given by \cite{Guoshen2010}
\begin{equation}
\mathbf{D}_{N \times GN} = \Big[ \mathbf{V}_1 \ldots \mathbf{V}_G \Big], ~ \mathbf{\Sigma}_g = \mathbf{V}_g  \mathbf{\Lambda}_g \mathbf{V}^T_g,~ g \in \{1, \ldots, G\},
\label{Eq2}
\end{equation}
based on the PCAs of the covariance matrices $\mathbf{\Sigma}_g$. In this dictionary, $\mathbf{x}$ is represented as,
\begin{equation}
\mathbf{x} = \mathbf{D} \boldsymbol{\alpha} = \Big[ \mathbf{V}_1 \ldots \mathbf{V}_G \Big] 
									~ \Big[ \mathbf{0}^T \ldots \boldsymbol{\alpha}^T_g \ldots \mathbf{0}^T \Big]^T 
									= \mathbf{V}_g  \boldsymbol{\alpha}_g,
\label{Eq3}
\end{equation}
\noindent
where $\| \boldsymbol{\alpha} \|_0 = \| \boldsymbol{\alpha}_g \|_0 = L \leq N \ll GN$, and $L$ corresponds to the largest eigenvalues of $\mathbf{\Sigma}_g$. Hence, the signal representation can be very sparse in this overcomplete structured dictionary, where typically $G \gg 1$. 

Now, let $\mathbf{y = \Phi x}  + \boldsymbol{\eta}$ be the CS signal, where $\mathbf{\Phi}_{M \times N}, M\ll N$ is a given (fixed for now) sensing matrix, and $\boldsymbol{\eta}$ is additive zero-mean white Gaussian noise, $\boldsymbol{\eta} \sim \mathcal{N} (\mathbf{0}, \sigma^2 \mathbf{I})$. Following an adequate initialization of the PCA basis \cite{Guoshen2010} for the signals at hand, the signal can be reconstructed from its projections using an iterative maximum \emph{a posteriori} based Expectation-Maximization (MAP-EM) algorithm, which simultaneously learns the GMM. In the E-step, a MAP estimate of the signal and the Gaussian (model) is obtained, and in the M-step, the parameters of the Gaussian are updated. The first part of the E-step corresponds to the MAP estimate of the signal, considering the Gaussian model $\mathbf{x} \sim \mathcal{N} (\mathbf{0}, \boldsymbol{\Sigma}_g)$ \cite{Guoshen2010},

\begin{equation}
\boldsymbol{\hat{\alpha}}_g = \mbox{argmin}_{ \boldsymbol{\alpha}_g}  \Big \{   \| \mathbf{y - \Phi \mathbf{V}}_g \boldsymbol{\alpha}_g  \|^2_2  
					   +  \sigma^2  \boldsymbol{\alpha}^T_g \mathbf{\Lambda}^{-1}_g \boldsymbol{\alpha}_g \Big \}.
\label{Eq4}
\end{equation}

\noindent Notice that we assumed here zero mean signals, which can be achieved simply by subtracting the Gaussian mean. Equation (\ref{Eq4}) can be efficiently solved in closed form using the Wiener filter $\mathbf{W}_g$, for each Gaussian $g \in \{1, \ldots, G\}$ {\cite{Guoshen2010},}
\begin{equation}
\boldsymbol{\hat{\alpha}}_g = \mathbf{W}_g \mathbf{y}, ~ 
\mathbf{W}_g = \boldsymbol{\Sigma}_g \mathbf{V}^T_g \mathbf{\Phi}^T \Big ( \mathbf{\Phi} \mathbf{V}_g \boldsymbol{\Sigma}_g \mathbf{V}^T_g \mathbf{\Phi}^T 
											+ \sigma^2 \mathbf{I}_M \Big ) ^{-1}.
\label{Eq5}
\end{equation}

\noindent Computing $\boldsymbol{\hat{\alpha}}_g$ for each $g \in \{1, \ldots, G\}$, the best Gaussian ($g$) (model selection) can be found at this step and from there an estimate of the signal $\mathbf{\hat{x}} = \mathbf{V}_g \boldsymbol{\hat{\alpha}}_g$, using (\ref{Eq5}).  

In the M-step, the Gaussian parameters are updated (see \cite{Guoshen2010} for a discussion on optimality of this),
\begin{equation}
\boldsymbol{\mu}_g = \frac{1}{| S_g |} \sum_{ i \in S_g } \mathbf{x}_i, ~ ~
\mathbf{\Sigma}_g = \frac{1}{| S_g |} \sum_{ i \in S_g } ( \mathbf{x}_i - \boldsymbol{\mu}_g ) ( \mathbf{x}_i - \boldsymbol{\mu}_g )^T  ,
\label{Eq6}
\end{equation}

\noindent where $S_g$ is the set of indices corresponding to the signals modeled best by the Gaussian $g$. Once we have updated the Gaussian parameters, the PCA decomposition of the covariance matrices (\ref{Eq2}) updates the dictionary. Usually, two iterations of the MAP-EM algorithm are enough for a given {\it fixed} $\mathbf{\Phi}$.

\section{Non-Adaptive (Batch) Statistical Compressive Sensing}
\label{sec_nonAdaptive}
A deterministic non-adaptive sensing matrix $\mathbf{\Phi}$ for SCS of GMMs can be used for both batch processing and adaptive SCS (ASCS). Indeed, a deterministic non-adaptive sensing matrix could be employed in the first step of the two-step adaptive SCS (Section \ref{sec_adaptive}), replacing the standard non-adaptive randomly constituted sensing matrix. As indicated in \cite{Duarte2012}, a deterministic non-adaptive sensing matrix, optimal for unstructured dictionaries \cite{Duarte2009}, already improves reconstruction with respect to a randomly constituted sensing matrix in SCS. However, using the optimal non-adaptive sensing matrix for structured dictionaries, proposed in \cite{Yonina}, does not achieve better results in this case \cite{Duarte2012}. The reason is that these sensing matrices do not exploit the known dictionary structure (Equation (\ref{Eq2})), nor the GMM employed.  The sensing approach introduced next does exploit these important properties.

We first encourage the RIP property by requiring that the columns of the equivalent dictionary $\mathbf{\Phi D}$ be as orthogonal as possible \cite{Duarte2009, Yonina}, 
\begin{equation}
\mathbf{\Phi} = \mbox{argmin}_{ \mathbf{ \hat{\Phi} } } \Big\{  \| \mathbf{D}^T \mathbf{ \hat{\Phi} }^T \mathbf{\hat{\Phi} D} - \mathbf{I}_{GN} \|^2_F \Big\}.
\label{Eq7}
\end{equation}

As shown in \cite{Yonina},
\begin{equation}
\| \mathbf{D}^T \mathbf{ \Phi }^T \mathbf{\Phi D} - \mathbf{I}_{GN} \|^2_F 
	= \|  \mathbf{ \Phi D} \mathbf{D}^T \mathbf{\Phi} ^T - \mathbf{I}_M \|^2_F + GN-M.
\label{Eq8}
\end{equation}
\noindent Since in CS, {$GN-M > 0$}, the minimum can be achieved by minimizing $\|  \mathbf{ \Phi D} \mathbf{D}^T \mathbf{\Phi} ^T - \mathbf{I}_M \|^2_F$. For the case of our SCS (Equation (\ref{Eq2})), 
\begin{equation}
\|  \mathbf{ \Phi D} \mathbf{D}^T \mathbf{\Phi} ^T - \mathbf{I}_M \|^2_F 
  	= \Big \|  \mathbf{\Phi} \big ( \sum^G_{g=1} \mathbf{V}_g \mathbf{V}^T_g \big ) \mathbf{\Phi}^T -  \mathbf{I}_M \Big \|^2_F 
	= \|  G \mathbf{ \Phi} \mathbf{\Phi} ^T - \mathbf{I}_M \|^2_F.
\label{Eq9}
\end{equation}

Hence, in order to encourage an RIP-type property, our SCS model only requires that the {\it rows} of the CS matrix be orthogonal, i.e., {$\mathbf{\Phi} \mathbf{\Phi}^T = \frac{1}{G} \mathbf{I}_M$. Since, $\frac{1}{G}$ is just a normalization constant, from now on and without any loss of generality, we will refer to this type of RIP condition as $\mathbf{\Phi} \mathbf{\Phi}^T = \mathbf{I}_M$.}

Since there are infinitely many matrices of size $M$$\times$$N$ with orthogonal rows, we are free to impose additional constrains to the CS matrices. It is well-known \cite{Guoshen2010, Guoshen2011} that for a Gaussian signal, $\mathbf{x} \sim \mathcal{N} (\boldsymbol{\mu}_g, \boldsymbol{\Sigma}_g)$, if we choose $\mathbf{\Phi} = [  \mathbf{v}^1_g \ldots \mathbf{v}^L_g ]^T$, where $\mathbf{v}^i_g, i \in \{1, \ldots, L\}$, are the first $L$ eigenvectors of the covariance matrix, then the linear estimate
\begin{equation}
\mathbf{\hat{x}} = \mathbf{\Phi}^T ( \mathbf{\Phi} \mathbf{x} )
= \sum_{i=1}^L \langle \mathbf{x} , \mathbf{v}^i_g \rangle \mathbf{v}^i_g
= \sum_{i=1}^L \lambda_i \mathbf{v}^i_g,    
\label{Eq10}
\end{equation}

\noindent minimizes the mean square reconstruction error (MSE, we will later deal with other tasks as well),
\begin{equation}
MSE = \| \mathbf{x} -\mathbf{\hat{x}} \|^2_2
= \| \mathbf{x} - \sum_{i=1}^L \langle \mathbf{x} , \mathbf{v}^i_g \rangle \mathbf{v}^i_g  \|^2_2
= \sum_{i=L+1}^N \lambda_i^2 ,
\label{Eq11}
\end{equation}

\noindent where $\lambda_i$ correspond to the $i$-th eigenvalue of the covariance matrix of the $g$-th Gaussian. Hence, for a given Gaussian $g$, we have
\begin{equation}
\mathbf{\Phi} \mathbf{V}_g = [  \mathbf{I}_M \quad \mathbf{0}_{M,N-M} ].
\label{Eq12}
\end{equation}

Given that in non-adaptive CS (or in the first step of the two-step adaptive CS), we do not know \emph{a priori} the corresponding Gaussian model of a given signal, {we could try to satisfy (\ref{Eq12}) on {\it average}} as
\begin{equation}
\mathbf{\Phi} = \mbox{argmin}_{ \mathbf{ \hat{\Phi} } } \Big\{ \Big \| \mathbf{ \hat{\Phi}} \sum^G_{g=1} p(g) \mathbf{V}_g - [  \mathbf{I}_m \quad \mathbf{0}_{M,N-M} ] \Big \|^2_F \Big\},  ~ s.t. ~ \mathbf{\hat{\Phi} \hat{\Phi}}^T = \mathbf{I}_M,
\label{Eq13}
\end{equation}
\noindent where we have also imposed the ``RIP'' (orthogonality) condition (\ref{Eq9}). {Another possibility could be to sense for the {\it average} Gaussian, this possibility is addressed in Section \ref{subsec_Classification}.} 

Let us define $\mathbf{\Phi} = [ \mathbf{I}_M \quad \mathbf{0}_{M, N-M} ] \mathbf{B}$, where $\mathbf{B}$ is an orthonormal basis in $\mathbb{R}^N$, and $\mathbf{E} = \sum^G_{g=1} p(g) \mathbf{V}_g$  is the expected Gaussian basis. Notice that since $\mathbf{B}$ is orthonormal, the rows of $\mathbf{\Phi}$ are orthonormal too, satisfying the condition (\ref{Eq9}). Equation (\ref{Eq13}) then simplifies to
\begin{align}
\mathbf{B} &= \mbox{argmin}_{ \mathbf{ \hat{B} } } \Big\{ \Big \| [  \mathbf{I}_M \quad \mathbf{0}_{M,N-M} ] \mathbf{ \hat{B} E} - [  \mathbf{I}_M \quad \mathbf{0}_{M,N-M} ] \Big \|^2_F \Big\}, ~ s.t. ~ \mathbf{\hat{B} \hat{B}}^T = \mathbf{I}_N \nonumber \\
 		&= \mbox{argmin}_{ \mathbf{ \hat{B} } } \Big\{ \Big \|  \mathbf{ \hat{B} E} - \mathbf{I}_N \Big \|^2_F \Big\}
~ s.t. ~ \mathbf{\hat{B} \hat{B}}^T = \mathbf{I}_N.
\label{Eq14}
\end{align}

\noindent The solution to (\ref{Eq14}) corresponds to a particular case of the generalized orthogonal  Procrustes problem \cite{Procrustes}, which in our case reduces to $\mathbf{B} = \mathbf{WU}^T$ (see derivation steps in Appendix \ref{App1}), where $\mathbf{E} = \mathbf{U \Delta W}^T$ is the singular value decomposition of $\mathbf{E}$. Hence, the solution to (\ref{Eq13}) is given by
\begin{equation}
\mathbf{\Phi} = [ \mathbf{I}_M \quad \mathbf{0}_{M,N-M} ] \mathbf{WU}^T,
\label{Eq15}
\end{equation}

\noindent i.e., the first $M$ rows of $\mathbf{WU}^T$ (see comments on computational complexity in the supplementary material). {We will denote the sensing matrix given in (\ref{Eq15}) as RIP-average of basis (RIP-AB).}  

To conclude, in this section, we proposed a new non-adaptive batch CS matrix for GMMs that can also be used in the first step of the proposed two-step adaptive framework,  as explained next.  

\section{Adaptive Task Driven Statistical Compressive Sensing}
\label{sec_adaptive}
As indicated in the introduction, the proposed two-step adaptive SCS algorithm  uses $K \le M \ll N$ measurements in the first step to identify the best Gaussian model for a given signal $\mathbf{x}$, and in the second step, it adds $M-K$ measurements using an adaptive or non-adaptive sensing matrix for that detected Gaussian. Several configurations of this two-step adaptive SCS are possible, with different degrees of adaptivity, as indicated in Table \ref{Table1}. {The corresponding computational complexity (see derivation details in the supplementary material) is also shown in this table, where $S$ is the number of signals, $\kappa$ the number of MAP-EM iterations (Section \ref{sec_SCS}), and $\chi$ the number of steepest ascent iterations (Section \ref{subsec_Classification}).} 

In the first step, we can use $K$ non-adaptive measurements, which can be random or the non-adaptive RIP-AB sensing matrix derived in Section \ref{sec_nonAdaptive}. Given that in the first step we are interested in feature selection for classification purposes  (detecting the Gaussian), we could also use non-adaptive information discriminant analysis (IDA) \cite{Nenadic} (see Section \ref{subsec_Classification}). Optionally, in the first step, $K$ adaptive measurements can be made, based on our extension of IDA, called here adaptive information discriminant analysis (AIDA) (Section \ref{subsec_Classification}). The number of adaptive measurements in the first step ($K$) can be automatically determined using sequential hypothesis testing (SHT), as decribed in Section \ref{subsec_SHT}. In the second step, the first $M-K$ eigenvalues of the corresponding covariance matrix (the Gaussian model has been identified in the first step) can be used, which as indicated before (Sections \ref{Introduction} and \ref{sec_nonAdaptive}) are optimal (in the MSE sense) for that Gaussian. However, this optimal sensing matrix disregards all previous $K$ measurements made in the first step.  Hence, we also provide here an optimal adaptive sensing matrix for the second step (reconstruction) that maximizes the mutual information (MI) between the CS signals and the (unknown) original signals, given that we now know the Gaussian model (estimated in the first step) and also the previous measurements (see also \cite{Carin2012}).  

Each of the possibilities in Table \ref{Table1}, with the exception of AIDA-SHT in the first step and the adaptive optimal (in the MI sense) sensing matrix in the second step, have already been defined (IDA is a particular case of AIDA). Hence, it remains only to introduce these new fully adaptive sensing protocols, which is the subject of the next two sections.  

\begin{table}
\caption{Two-steps Adaptive Statistical CS configurations.}
\begin{center}
\tiny
\begin{tabular}{| c | c | c | c | c |}
\hline 
\multicolumn{2}{|c|}{\textbf{Step 1. Classification/Detection (unknown Gaussian)}} & \multicolumn{2}{c|}{\textbf{Step 2. Reconstruction (known Gaussian)}} & \textbf{Computational Complexity}\\ \cline{1-4} 
\textbf{Sensing} & \textbf{Adaptivity} & \textbf{Sensing} & \textbf{Adaptivity} & (see details in the supplementary material)\\ \cline{1-5}

Random	& No &  Optimal (MSE), non-adaptive (Section \ref{sec_nonAdaptive}) &	No & $O(\kappa SGMN^2)$\\
RIP-AB (Section \ref{sec_nonAdaptive}) & No & Optimal (MSE), non-adaptive (Section \ref{sec_nonAdaptive}) & No & $O(\kappa G(N^3 + SMN))$\\
IDA (Section \ref{subsec_Classification}) & No & Optimal (MSE), non-adaptive (Section \ref{sec_nonAdaptive}) & No & $O(\kappa (\chi M^3 + GSMN))$\\
IDA (Section \ref{subsec_Classification}) & No & Optimal (MI), adaptive (Section \ref{subsec_Reconstruction}) & Yes & $O(\kappa (\chi M^3 + GSMN))$\\
AIDA-SHT (Sections \ref{subsec_Classification}, \ref{subsec_SHT}) & Yes & Optimal (MI), adaptive (Section \ref{subsec_Reconstruction}) & Yes & 
$O(\kappa S (\chi M^4 + GMN^2))$ \\
\hline
\end{tabular}
\end{center}
\label{Table1}
\end{table}

Note that there are other configurations possible in this two-step framework that are not in Table \ref{Table1}. For instance, using the optimal (MI) adaptive sensing matrix in the second step, when in the first step non-adaptive random or RIP-AB measurements were made, or using AIDA with a pre-defined number of steps (without SHT). However, we limit ourselves here to the indicated configurations, since these are the most representative, and from these configurations one can infer the behavior of others. This great deal of flexibility of the proposed two-step adaptive SCS framework makes it  very attractive, providing different options to choose from depending on the application and available computational resources. The most expensive computational costs incurred by these configurations (indicated in Table \ref{Table1}) can be done offline, before the actual sensing takes place, as detailed in the complexity analysis in the supplementary material.

\subsection{Step 1: Classification/Detection}
\label{subsec_Classification}
We extend the information discriminant analysis (IDA) framework \cite{Nenadic}, that performs non-adaptive linear measurements aimed at extracting the best features for classification, to an adaptive IDA (AIDA). Let $\mathbf{y}_{(k-1)} = [\mathbf{y}_1 ~\ldots~ \mathbf{y}_{k-1}]$, be the previous (known) $k-1$ measurements of size $b \ge 1$ each, and $\mathbf{y}_k$ the (unknown) next $k$-th measurements of size $b$. Also, let $\mathbf{y}_{(k-1)} = \mathbf{\Phi}_{(k-1)} \mathbf{x}$, where $\mathbf{\Phi}_{(k-1)} = [\mathbf{\Phi}^T_1 ~\ldots~ \mathbf{\Phi}^T_{k-1}]^T$ is the $(k-1)b$$\times$$N$ (known) sensing matrix used so far, and $\mathbf{\Phi}_k$ the $b$$\times$$N$ (unknown) sensing matrix for the next $b$ measurements, i.e., $\mathbf{y}_k = \mathbf{\Phi}_k \mathbf{x}$. Note that consistent with this notation, $\mathbf{y}_{(k)} = \mathbf{\Phi}_{(k)} \mathbf{x}$, which corresponds to all measurements made so far plus the new ones.

We want to find $\mathbf{\Phi}_k$ that maximizes the mutual information (MI) between the new measurements and the unknown Gaussian class $g \in \{0, \ldots, G\}$, given all previous measurements, $I(\mathbf{y}_k ; g | \mathbf{y}_{(k-1)})$,
\begin{equation}
\mathbf{\Phi}_k = \mbox{argmax}_{\mathbf{\hat{\Phi}}_k} I(\mathbf{y}_k ; g | \mathbf{y}_{(k-1)}), 
~s.t.~ \mathbf{\hat{\Phi}}_k \mathbf{\hat{\Phi}}^T_k = \mathbf{I}_b,
\label{Eq16}
\end{equation}

\noindent where we have also imposed orthonormality following the RIP condition (see Section \ref{sec_nonAdaptive}), also constraining the amount of energy required by the sensor \cite{Carin2012, Nenadic, Neifeld2009}. By definition of the conditional mutual information \cite{MI_book},
\begin{equation}
I(\mathbf{y}_{k} ; g | \mathbf{y}_{(k-1)}) = E_{\mathbf{y}_{k}, \, g, \, \mathbf{y}_{(k-1)}} \Big \{ \log \frac{ p( \mathbf{y}_k, g | \mathbf{y}_{(k-1)} ) } { p( \mathbf{y}_k | \mathbf{y}_{(k-1)} ) p(g | \mathbf{y}_{(k-1)} ) }  \Big \},
\label{Eq17}
\end{equation}

\noindent where $E\{ \centerdot \}$ represents expectation. From now on, and for simplicity, we omit the dependent variables in the expectation. 

Using Baye's rule and logarithm properties, (\ref{Eq17}) can be expanded to (see details in Appendix \ref{App2}), 
\begin{equation}
 I(\mathbf{y}_{k} ; g | \mathbf{y}_{(k-1)}) = H \big( p( \mathbf{y}_{(k)} ) \big) - H \big( p( \mathbf{y}_{(k)} | g) \big) 
 	- \Big[ H \big( p( \mathbf{y}_{(k-1)} ) \big) - H \big( p( \mathbf{y}_{(k-1)} | g ) \big) \Big ],
 \label{Eq18}
\end{equation}  

\noindent where $H( \centerdot )$ is the differential entropy. Since the entropies within brackets do not depend on $\mathbf{\Phi}_k$, we consider them constants and denoted by $C(\mathbf{y}_{(k-1)})$. Given that $\mathbf{y}_{(k)} = \mathbf{\Phi}_{(k)} \mathbf{x}$ is a linear transformation of a Gaussian ($g \in \{1, \ldots,G\}$) random vector $\mathbf{x}$, the class conditional probability is given by \cite{ProbabilityBook}
\begin{equation}
 p( \mathbf{y}_{(k)} | g) = \frac{1}{ (2 \pi)^{kb/2} \sqrt { |  \mathbf{\Sigma}_{\mathbf{y}_{(k)} | g}  | } } 
	exp \Big ( -\frac{1}{2} (\mathbf{y}_{(k)} - \boldsymbol{\mu}_{\mathbf{y}_{(k)} | g})^T \mathbf{\Sigma}^{-1}_{\mathbf{y}_{(k)} | g} (\mathbf{y}_{(k)} - \boldsymbol{\mu}_{\mathbf{y}_{(k)} | g}) \Big ),  
 \label{Eq19}
\end{equation}

\noindent where $\mathbf{\Sigma}_{\mathbf{y}_{(k)} | g} = \mathbf{\Phi}_{(k)} \mathbf{\Sigma}_g \mathbf{\Phi}^T_{(k)} + \sigma^2 \mathbf{I}$ and $\boldsymbol{\mu}_{\mathbf{y}_{(k)} | g} = \mathbf{\Phi}_{(k)}  \boldsymbol{\mu}_g$. The entropy of $p( \mathbf{y}_{(k)} | g)$ (see (\ref{Eq18})) has a known closed-form given by \cite{MI_book, Mukund}
\begin{equation}
H \big ( p( \mathbf{y}_{(k)} | g) \big) =  \frac{1}{2} \Big [kb(1 + \log(2 \pi)) + \sum^G_{g=1} p(g) \log | \mathbf{\Sigma}_{\mathbf{y}_{(k)} | g}  |  \Big ],
\label{Eq20}
\end{equation}

\noindent where as before {$p(g)$ is the probability that $\mathbf{x} \sim \mathcal{N} (\boldsymbol{\mu}_g, \boldsymbol{\Sigma}_g)$}. 
 
On the other hand, the entropy of $p( \mathbf{y}_{(k)} )$, 
\begin{equation}
H \big ( p( \mathbf{y}_{(k)} ) \big ) =  - \int p( \mathbf{y}_{(k)} ) \log ~ p( \mathbf{y}_{(k)} ) d \mathbf{y}_{(k)}, 
\label{Eq21}
\end{equation}

\noindent in (\ref{Eq18}) has no closed-form. Rather than numerically computing (\ref{Eq21}), we use here the approximation in \cite{Nenadic, Mukund}, and define
\begin{eqnarray}
\tilde{p}( \mathbf{y}_{(k)}) = \frac{1}{ (2 \pi)^{kb/2} \sqrt { |  \mathbf{\bar{\Sigma}}_{\mathbf{y}_{(k)}}  | } }  exp \Big ( -\frac{1}{2} (\mathbf{y}_{(k)} - \boldsymbol{\bar{\mu}}_{\mathbf{y}_{(k)}})^T \mathbf{\bar{\Sigma}}^{-1}_{\mathbf{y}_{(k)}} (\mathbf{y}_{(k)} - \boldsymbol{\bar{\mu}}_{\mathbf{y}_{(k)}}) \Big ),  \\
	\mathbf{\bar{\Sigma}}_{\mathbf{y}_{(k)}} = \sum^G_{g=1} p(g) \big[ \mathbf{\Sigma}_{\mathbf{y}_{(k)} | g} 
	+ (\boldsymbol{\bar{\mu}}_{\mathbf{y}_{(k)}} - \boldsymbol{\mu}_{\mathbf{y}_{(k)} | g}) (\boldsymbol{\bar{\mu}}_{\mathbf{y}_{(k)}} - \boldsymbol{\mu}_{\mathbf{y}_{(k)} | g})^T \big], ~  \boldsymbol{\bar{\mu}}_{\mathbf{y}_{(k)}} = \sum^G_{g=1} p(g) \boldsymbol{\mu}_{\mathbf{y}_{(k)} | g}, \nonumber 	
\label{Eq22}
\end{eqnarray}

\noindent whose entropy has closed-form and provides an upper bound to the entropy of $p( \mathbf{y}_{(k)} )$ \cite{Mukund}, i.e.,
\begin{equation}
H \big ( p( \mathbf{y}_{(k)} ) \big ) \le  - \int \tilde{p}( \mathbf{y}_{(k)} ) \log ~ \tilde{p}( \mathbf{y}_{(k)} ) d \mathbf{y}_{(k)} =  \frac{1}{2} \big [ kb (1 + \log(2 \pi)) + \log | \mathbf{\bar{\Sigma}}_{\mathbf{y}_{(k)}} | \big ].
\label{Eq23}
\end{equation} 

Since in our statistical CS model (Section \ref{sec_SCS}), we made $\boldsymbol{\mu}_{\mathbf{y}_{(k)} | g} = \mathbf{0}$, hence, $\boldsymbol{\bar{\mu}}_{\mathbf{y}_{(k)}} = \mathbf{0}$ and
\begin{equation}
\mathbf{\bar{\Sigma}}_{\mathbf{y}_{(k)}} = \mathbf{\Phi}_{(k)} \mathbf{\bar{\Sigma}} \mathbf{\Phi}^T_{(k)} + \sigma^2 \mathbf{I}, ~ \mathbf{\bar{\Sigma}} =  \sum^G_{g=1} p(g) \mathbf{\Sigma}_g.
\label{Eq24}
\end{equation}

\noindent Replacing (\ref{Eq20}) and (\ref{Eq23}) in (\ref{Eq18}), we obtain 
\begin{equation}
 I(\mathbf{y}_{k} ; g | \mathbf{y}_{(k-1)}) \le  
 \frac{1}{2} \Big ( \log | \mathbf{\bar{\Sigma}}_{\mathbf{y}_{(k)}} | - \sum^G_{g=1} p(g) \log | \mathbf{\Sigma}_{\mathbf{y}_{(k)} | g}  | \Big )
  + C(\mathbf{y}_{(k-1)}) = \mu (\mathbf{y}_k ; g | \mathbf{y}_{(k-1)}) ,
 \label{Eq25}
\end{equation}  

\noindent where as in IDA \cite{Nenadic}, we have defined the right term as an $\mu$-measure, providing a more mathematically tractable problem and also a general measure of class discrimination that extends beyond Gaussian distributions. {Hence, instead of maximizing the MI, which has no closed form, we propose to maximize the $\mu$-measure, given by}   
\begin{equation}
\mathbf{\Phi}_k = \mbox{argmax}_{\mathbf{\hat{\Phi}}_k} \mu(\mathbf{y}_{k} ; g | \mathbf{y}_{(k-1)}),
~s.t.~ \mathbf{\hat{\Phi}}_k \mathbf{\hat{\Phi}}^T_k = \mathbf{I}_b.
\label{Eq26}
\end{equation}

{As shown in \cite{Nenadic}, $\mu$ is a class-separability measure that is optimal in the Bayes sense, when the noise is uncorrelated to the classes and the class differences are all in the signal subspace}. After some mathematical derivations (see details in Appendix \ref{App3}), the adaptive $\mu$-measure (\ref{Eq26}) can be rewritten as
\begin{align}
 \mu (\mathbf{y}_{k} ; g | \mathbf{y}_{(k-1)}) &=  
 \frac{1}{2} \Big ( \log | \mathbf{\Phi}_k \mathbf{\bar{P}} \mathbf{\Phi}^T_k | 
- \sum^G_{g=1} p(g) \log | \mathbf{\Phi}_k \mathbf{P}_g \mathbf{\Phi}^T_k | \Big ) + D(\mathbf{y}_{(k-1)}),  
\label{Eq27} \\
 \mathbf{P}_g &= \mathbf{\Sigma}_g - \mathbf{\Sigma}_g \mathbf{\Phi}^T_{(k-1)} \mathbf{\Sigma}^{-1}_{\mathbf{y}_{(k-1)}|g} \mathbf{\Phi}_{(k-1)} \mathbf{\Sigma}_g + \sigma^2 \mathbf{I}_b, \nonumber   \\
\mathbf{\bar{P}} &= \mathbf{\bar{\Sigma}} - \mathbf{\bar{\Sigma}} \mathbf{\Phi}^T_{(k-1)} \mathbf{\bar{\Sigma}}^{-1}_{\mathbf{y}_{(k-1)}} \mathbf{\Phi}_{(k-1)} \mathbf{\bar{\Sigma}} + \sigma^2 \mathbf{I}_b,   \nonumber  
\end{align}  

\noindent where $D(\mathbf{y}_{(k-1)})$ accounts for all the terms involving previous measurements which do not depend on $\mathbf{\Phi}_{k}$. In (\ref{Eq27}), the adaptive $\mu$-measure depends not only on the (unknown) new block sensing $\mathbf{\Phi}_k$ matrix,\footnote{We consider the general case of block size $b \ge 1$. If $b=1$, then we have adaptation for each new measurement.} but also on the sensing matrix we have so far $\mathbf{\Phi}_{(k-1)}$. The solution to (\ref{Eq26}) can be obtained via steepest ascent, with gradient given by (see useful vector derivatives in \cite{MatrixCookbook, Nenadic})

\begin{equation}
\frac{\partial \mu(\mathbf{y}_{k} ; g | \mathbf{y}_{(k-1)})}{\partial \mathbf{\Phi}_k} = ( \mathbf{\Phi}_k \mathbf{\bar{P}} \mathbf{\Phi}^T_k)^{-1} \mathbf{\Phi}_k \mathbf{\bar{P}} - \sum^G_{g=1} p(g) ( \mathbf{\Phi}_k \mathbf{P}_g \mathbf{\Phi}^T_k)^{-1} \mathbf{\Phi}_k \mathbf{P}_g.
\label{Eq28}
\end{equation}

\noindent The condition $\mathbf{\Phi}_k \mathbf{\Phi}^T_k = \mathbf{I}_b$ in (\ref{Eq26}) can be imposed at the end of each steepest ascent iteration as in \cite{Nenadic}. 

Of course, for $k=1$, we do not have previous measurements, hence, we can use the non-adaptive IDA, which we repeat here for completeness of the presentation \cite{Nenadic},
\begin{equation}
 \mu (\mathbf{y}_{k} ; g) =  \frac{1}{2} \Big ( \log | \mathbf{\Phi}_1 \mathbf{\bar{\Sigma}} \mathbf{\Phi}^T_1 | 
- \sum^G_{g=1} p(g) \log | \mathbf{\Phi}_1 \mathbf{\Sigma}_g \mathbf{\Phi}^T_1 | \Big ).
\label{Eq29}
\end{equation}  

\noindent The maximum $\mu$-measure, conditioned by $\mathbf{\Phi}_1 \mathbf{\Phi}^T_1 = \mathbf{I}_b$, is obtained by steepest ascent with gradient \cite{Nenadic}
\begin{equation}
\frac{\partial \mu(\mathbf{y}_{1} ; g)}{\partial \mathbf{\Phi}_1} = ( \mathbf{\Phi}_1 \mathbf{\bar{\Sigma}} \mathbf{\Phi}^T_1)^{-1} \mathbf{\Phi}_1 \mathbf{\bar{\Sigma}} - \sum^G_{g=1} p(g) ( \mathbf{\Phi}_1 \mathbf{\Sigma}_g \mathbf{\Phi}^T_1)^{-1} \mathbf{\Phi}_1 \mathbf{\Sigma}_g.
\label{Eq30}
\end{equation}  

\noindent It is worth nothing here the similarity between the IDA equations (\ref{Eq29}) and (\ref{Eq30}), and AIDA equations (\ref{Eq27}) and (\ref{Eq28}).\footnote{The IDA equations presented here for $k=1$ are indeed the general IDA equations, since the block size $b$ can be of any size.} Indeed, the role that $\mathbf{\bar{\Sigma}}$ and $\mathbf{\Sigma}_g$ play on IDA corresponds to $\mathbf{\bar{P}}$ and $\mathbf{P}_g$, respectively, on AIDA. {Note also that IDA uses the average (expected) Gaussian ($\mathbf{\bar{\Sigma}}, \mathbf{\bar{\mu}=0}$).}  

After $k$ adaptive measurements, $\mathbf{y}_{(k)}$, the identification of the Gaussian (classification) can be performed using any {\it local} classifier. In particular, we can use the MAP criteria,
\begin{equation}
g = \mbox{argmin}_{\hat{g}} \Big \{ \mathbf{y}^T_{(k)} \mathbf{\Sigma}^{-1}_{\mathbf{y}_{(k)} | \hat{g}} \mathbf{y}_{(k)} 
+ \log | \mathbf{\Sigma}_{\mathbf{y}_{(k)} | \hat{g}} | \Big \},
\label{Eq31}
\end{equation}

\noindent for classification purposes (as used in Section \ref{subsec_expClass}), or the MAP criteria indicated in Section \ref{sec_SCS} (equations (\ref{Eq4}) and (\ref{Eq5})), that is based on $\mathbf{\hat{x}}$, which is better for reconstruction purposes (used in sections \ref{subsec_expBlockSCS}, \ref{subsec_expASCS}). 

\subsection{Step 2: Reconstruction}
\label{subsec_Reconstruction}

Having identified the Gaussian model for the signal in the first step, we focus now on finding adaptively the best reconstruction sensing matrix for this Gaussian (as mentioned before, this step can be skipped if the detected Gaussian/model is not of interest for reconstruction). One possibility is to use the optimal (in the MSE sense) sensing matrix for this Gaussian, as indicated in Section \ref{sec_nonAdaptive}. However, this approach disregards all the previous information ($K$ measurements), and hence, it is in general suboptimal. 

At this step, we want to maximize the MI between the signal $\mathbf{x}$ and the new measurements $\mathbf{y}_k$ of size $b$ (see also \cite{Carin2012}, where MI for kernel design is introduced and elegantly analyzed), having into account previous measurements $\mathbf{y}_{(k-1)}$ (including those from the first step and any other previous measurements at this second step), and also the now detected Gaussian, $\gamma \in \{1 \ldots G\}$, i.e., we want to find
\begin{equation}
\mathbf{\Phi}_k = \mbox{argmax}_{\mathbf{\hat{\Phi}}_k} I(\mathbf{y}_k; \mathbf{x}  | \mathbf{y}_{(k-1)}, \gamma), 
~s.t.~ \mathbf{\hat{\Phi}}_k \mathbf{\hat{\Phi}}^T_k = \mathbf{I}_b.
\label{Eq32}
\end{equation}

\noindent By the definition of conditional mutual information \cite{MI_book},
\begin{equation}
I(\mathbf{y}_k; \mathbf{x}  | \mathbf{y}_{(k-1)}, \gamma) = I(\mathbf{y}^\gamma_k; \mathbf{x}^\gamma  | \mathbf{y}^\gamma_{(k-1)})
= E \Big \{ \log \frac{ p( \mathbf{y}^\gamma_k, \mathbf{x}^\gamma | \mathbf{y}^\gamma_{(k-1)} ) } { p( \mathbf{y}^\gamma_k | \mathbf{y}^\gamma_{(k-1)} ) p(\mathbf{x}^\gamma | \mathbf{y}^\gamma_{(k-1)} ) }  \Big \},
\label{Eq33}
\end{equation}

\noindent where the knowledge of the Gaussian identity has been used to specify the signal $\mathbf{x}^\gamma \sim \mathcal{N} (\boldsymbol{\mu}_\gamma, \boldsymbol{\Sigma}_\gamma)$, previous measurements $\mathbf{y}^\gamma_{(k-1)} = \mathbf{\Phi}_{(k-1)} \mathbf{x} ^ \gamma$, and new (unknown) measurements $\mathbf{y}^\gamma_k = \mathbf{\Phi}_k \mathbf{x} ^ \gamma$. 

As detailed in Appendix \ref{App4}, (\ref{Eq32}) has a closed-form solution, given by
\begin{align}
\mathbf{\Phi}_k &= [ \mathbf{u}_1 ~ \ldots ~ \mathbf{u}_{M-K}]^T, ~ \mathbf{U} = [ \mathbf{u}_1 ~ \ldots ~ \mathbf{u}_{N}], 
\label{Eq34} \\
 \mathbf{P}_\gamma &= \mathbf{U} \mathbf{\Delta} \mathbf{U}^T, ~\mathbf{P}_\gamma =  \mathbf{\Sigma}_\gamma - \mathbf{\Sigma}_\gamma \mathbf{\Phi}^T_{(k-1)} \mathbf{\Sigma}^{-1}_{\mathbf{y}^\gamma_{(k-1)}} \mathbf{\Phi}_{(k-1)} \mathbf{\Sigma}_\gamma + \sigma^2 \mathbf{I}_b. \nonumber
\end{align}

\noindent Hence, the optimal (in the MI sense) CS matrix in the second step is given by the transposed first $M-K$ eigenvectors of the matrix $\mathbf{P}_\gamma$, which is the same as in (\ref{Eq27}), but now for a known Gaussian $\gamma$. Notice that if there are no previous measurements (non-adaptive CS), $\mathbf{P}_\gamma = \mathbf{\Sigma}_\gamma$, and the optimal (in the MI sense) CS matrix for a known Gaussian corresponds to the transposed first $M-K$ eigenvectors of the coresponding covariance matrix, which is exactly the optimal non-adaptive sensing matrix (in the MSE sense) given in Section \ref{sec_nonAdaptive}.   

\subsection{Sequential Hypothesis Testing}
\label{subsec_SHT}
We defined AIDA in Section \ref{subsec_Classification} for the first step, where the task is classification. However, the number of measurements required in this step, $K$, was assumed to be a known parameter. We use here the sequential hypothesis testing (SHT) framework in \cite{Neifeld2009} to automatically determine the number of measurements $K$ in the first step.  SHT has been shown \cite{SHT1, SHT2} to lead to fewer measurements on average, compared to hypothesis testing approaches that use a fixed number of measurements.
\begin{algorithm}                      
\caption{Sequential Hypothesis Testing (SHT)}         
\label{alg1}                           
\begin{algorithmic}    
\scriptsize                
\REQUIRE $P_e, ~G, ~M, ~b, ~p(g=1), \ldots, ~p(g=G), ~\mathbf{\Sigma}_1, ~\ldots, ~\mathbf{\Sigma}_G$  
\STATE $k = 1$
\STATE $\eta = \frac{1 - P_e}{P_e}$ \quad \{ Stopping threshold \}
\STATE $\mathbf{\Phi}_{1} = IDA(b, ~p(g=1), \ldots, ~p(g=G), ~\mathbf{\Sigma}_1, ~\ldots, ~\mathbf{\Sigma}_G$	\quad \{ Initialization \}
\WHILE[OPTIMIZATION]{$kb \le M$}
	\STATE $k = k+1$
	\STATE Initialize $\mathbf{\Phi}_k$ (usually with a random sensing matrix)
	
	\WHILE[Steepest Ascend]{ $\mu(\mathbf{y}_{k} ; g | \mathbf{y}_{(k-1)})$ increases (Equation (\ref{Eq27}))}
	\STATE $\mathbf{\Phi}_k \gets \mathbf{\Phi}_k + \alpha \frac{\partial \mu(\mathbf{y}_{k} ; g | \mathbf{y}_k)}{\partial \mathbf{\Phi}_k}$ 
			\quad \{ Using Equation (\ref{Eq28}) \}
	\ENDWHILE
	\STATE $\mathbf{\Phi}_{(k)} \gets [\mathbf{\Phi}^T_{(k-1)} ~ \mathbf{\Phi}^T_k]^T $, 	$\mathbf{y}_k \gets \mathbf{\Phi}_k \mathbf{x}$	\quad \{Updates CS matrix\} 
	
	\FOR[Bayesian Update of Priors]{$g = 1 \to G$} 
	\STATE $p(g) \gets \frac{p(\mathbf{y}_k | g) p(g)}{p(\mathbf{y}_k)} = \frac{p(\mathbf{y}_k | g) p(g)}{\sum^G_{i=1}p(\mathbf{y}_k | g=i) p(g=i)} $
	\ENDFOR
	
	\FOR[Likelihood Ratios]{$i, j = 1 \to G$}
		\STATE $L_{i,j} \gets \frac{\prod^k_{l=1} p(\mathbf{y}_{(l)} | g=i)}{\prod^k_{l=1} p(\mathbf{y}_{(l)} | g=j)}\frac{p(g=i)}{p(g=j)}$
	\ENDFOR
	
	\IF[Classification]{$\forall_{i \ne j} ~ L_{i,j} > \eta$}
		\STATE $\gamma \gets i$
		\RETURN $\gamma, \mathbf{\Phi}_k$
	\ENDIF
	
\ENDWHILE
\end{algorithmic}
\end{algorithm}

The main idea is to use SHT to obtain a minimum possible number of measurements $K$ in the first (classification) step, leaving as many samples ($M-K$) as possible for the second (reconstruction) step, where an optimal sensing can be performed given the detected Gaussian and all the previous measurements. It is clear, however, that an incorrect detection in the first step would lead to sub-optimal measurements in the second step. Hence, we need a way to decide when to stop, based on a given probability of classification error $P_e$,

\begin{equation}
P_e = \sum^G_{g=1} p(1, \ldots, g-1, g+1, \ldots, G)p(g), 
\end{equation}

\noindent where $p(1, \ldots, g-1, g+1, \ldots, G)$ represents the misclassification error probability, when $g$ is the true Gaussian. This is precisely what SHT provides, based on Bayesian updates of the prior class probabilities and maximum likelihood. Algorithm \ref{alg1} describes in detail the SHT method \cite{Neifeld2008b, Neifeld2009}.

We have provided an algorithm, based on hypothesis testing, to automatically determine the number of measurements required in the first step of the two-step framework, for a given probability of classification error. This in conjunction with AIDA, provides the AIDA-SHT CS protocol cited in Table \ref{Table1}. 

\section{Experimental Results}
\label{sec_experiments}
\subsection{Non-Adaptive Statistical Compressive Sensing}
\label{subsec_expBlockSCS}
We compare the reconstruction performance of the proposed non-adaptive sensing RIP-AB (Section \ref{sec_nonAdaptive}), with random sensing, the optimal non-adaptive sensing for unstructured dictionaries \cite{Duarte2009, Duarte2012}, and the optimal sensing for structured dictionaries \cite{Yonina}. {We use all 8$\times$8 overlapping patches from 20 natural images taken from the Berkeley segmentation dataset \cite{BerkelyDataSet} in order to (offline) adapt the learned dictionaries (GMM) and sensing matrices (which depend on the PCA basis of the GMM, except for random sensing of course) to each image. The (online) evaluation of the learned dictionaries and non-random sensing matrices was done using non-overlapping 8$\times$8 patches.} Only $\kappa$ = 11 iterations of the MAP-EM statistical CS were needed to learn the dictionaries and non-random sensing matrices.\footnote{We require more MAP-EM iterations than those indicated on Section \ref{sec_SCS}, since now, the sensing matrix $\mathbf{\Phi}$ and the dictionary (GMM) are both being simultaneously adapted.}

\begin{table}[h!b!p!]
\scriptsize
\caption{Mean Image Reconstruction PSNR (dbs), non-overlapping patches.}
\begin{center}
\begin{tabular}{| c | c | c | c | c |}
\hline
\textbf{Compression Ratio} & \multicolumn{4}{c}{\textbf{Sensing Matrix}} \vline\\
\cline{2-5}
                                   & \textbf{Random}   & \textbf{Optimal Unstructured} & \textbf{Optimal Structured} & \textbf{RIP-AB GMM}\\
\hline 
8.0 & 28.04 & 28.74 & 28.93 & 29.30\\
\hline
5.3 & 29.86 & 30.42 & 30.59 & 30.98\\
\hline
3.2 & 32.90 & 33.47 & 33.53 & 34.0\\
\hline
\end{tabular}
\end{center}
\label{Table2}
\end{table}

\begin{figure}[!hbp]
\begin{center}
\scriptsize
\includegraphics[width=0.48\textwidth, height=0.6\textwidth]{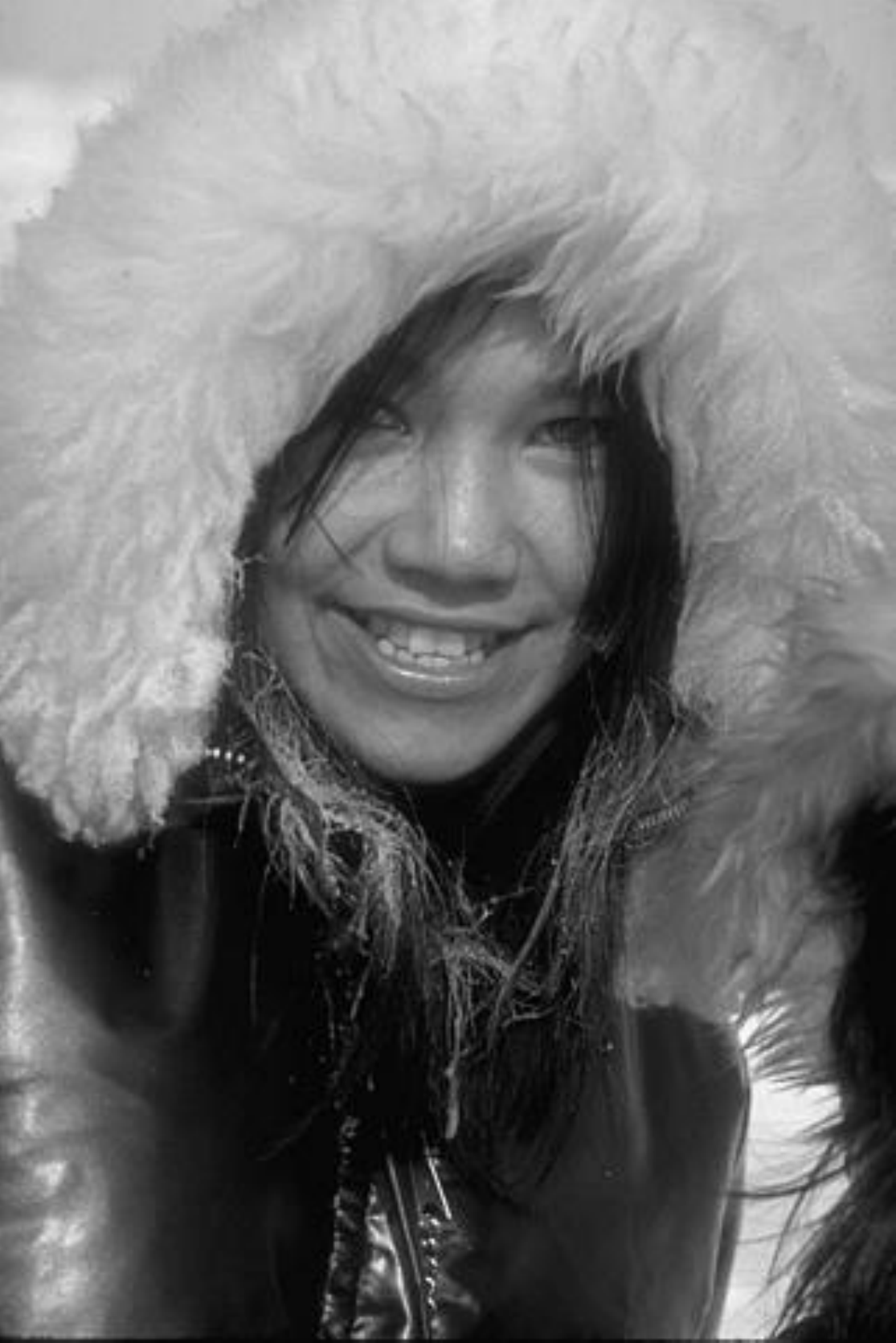}%
\quad \includegraphics[width=0.48\textwidth, height=0.6\textwidth]{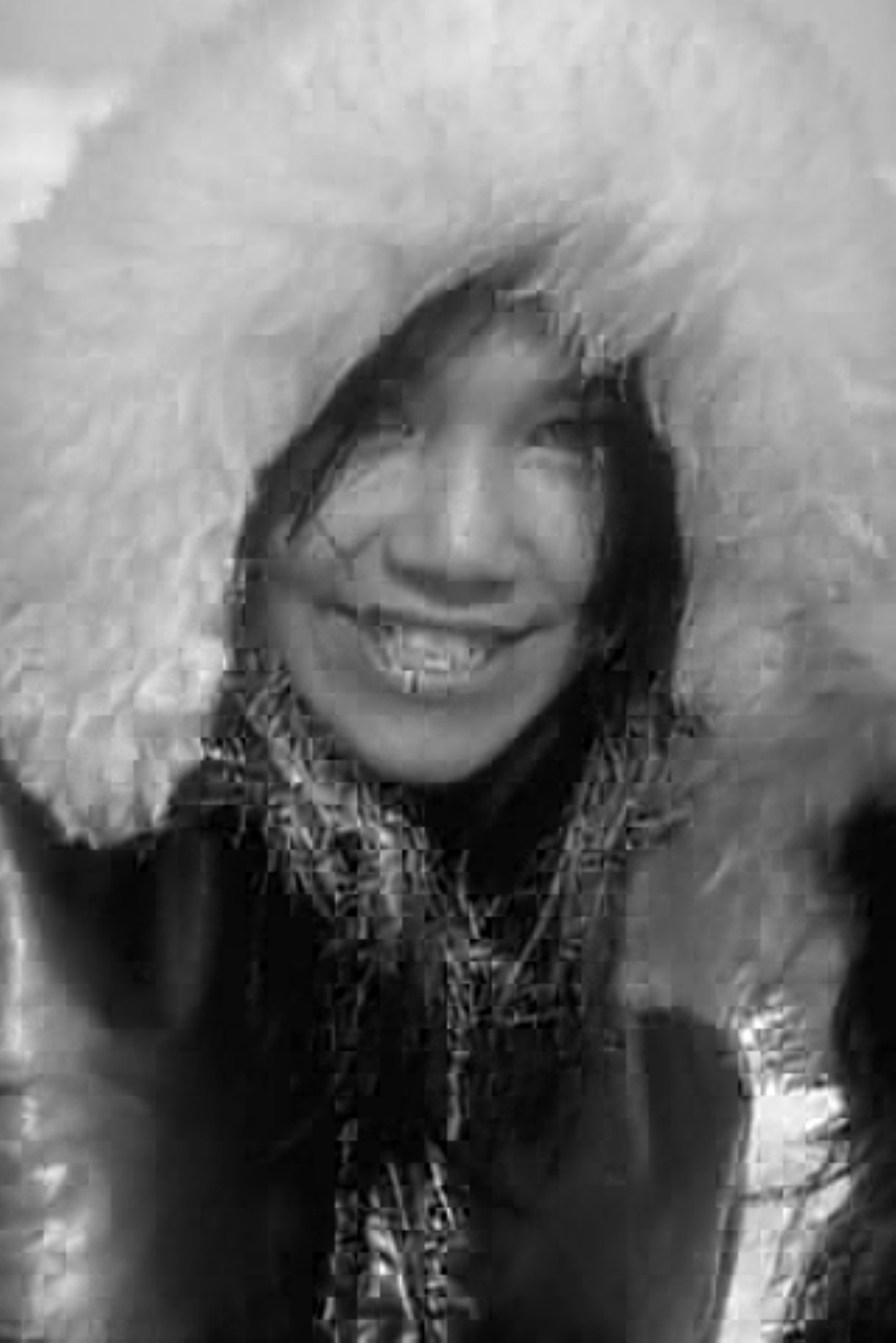}\\
(a) \qquad \qquad \qquad \qquad \qquad \qquad \qquad \qquad \qquad \qquad \qquad \qquad \qquad \qquad (b)\\
\includegraphics[width=0.48\textwidth, height=0.6\textwidth]{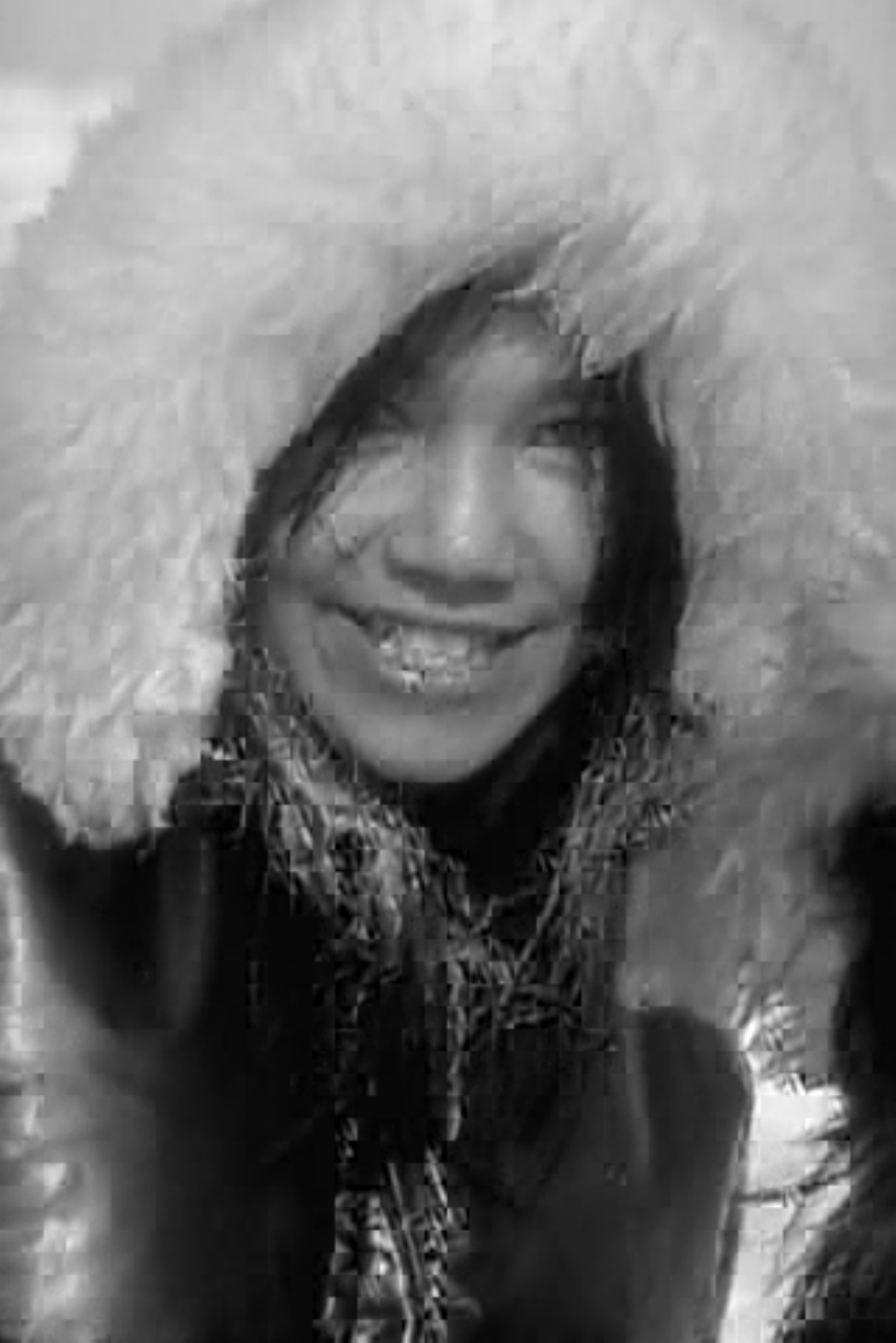}%
\quad \includegraphics[width=0.48\textwidth, height=0.6\textwidth]{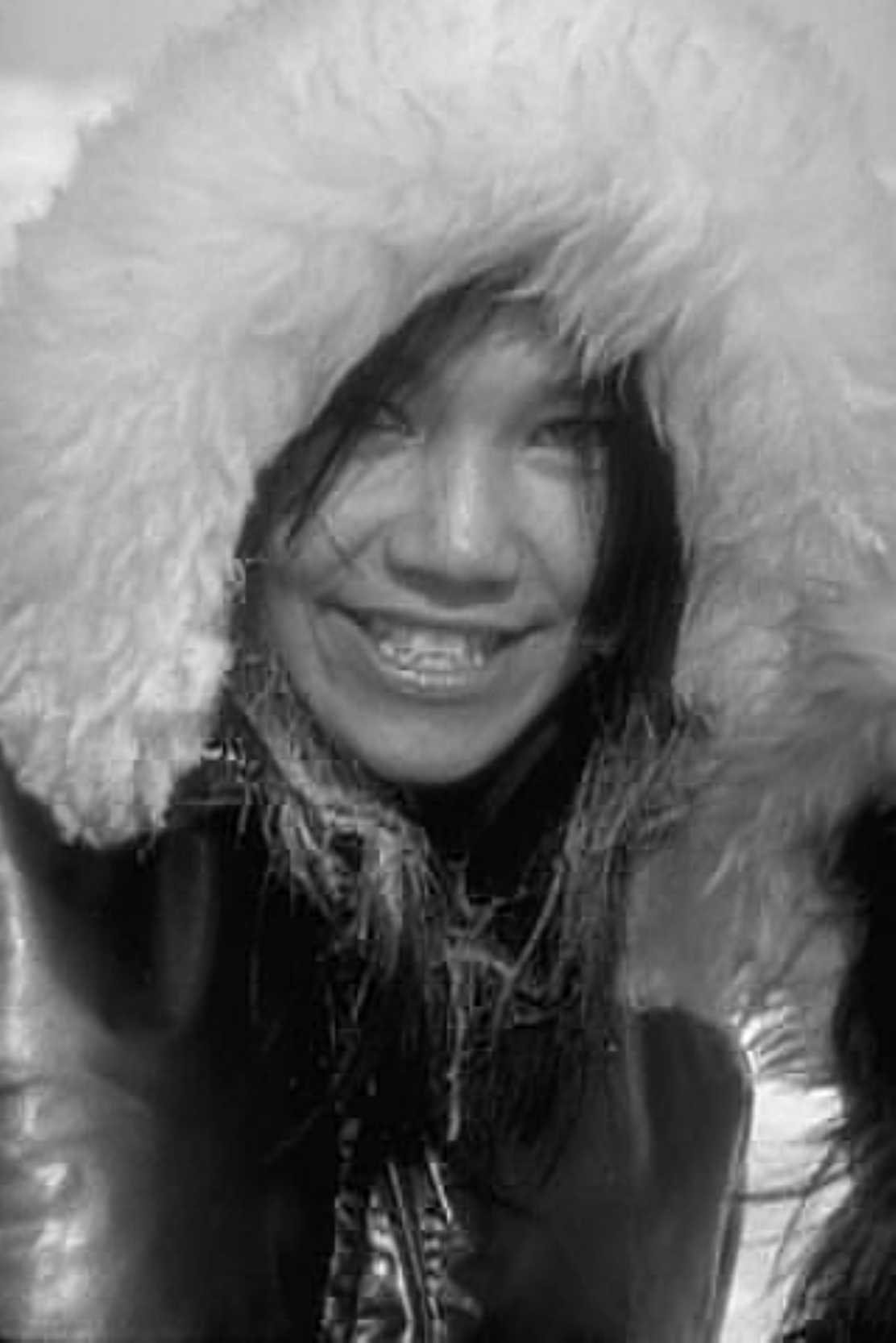}\\
(c) \qquad \qquad \qquad \qquad \qquad \qquad \qquad \qquad \qquad \qquad \qquad \qquad \qquad \qquad (d)
\end{center}
\caption{Reconstructed image from learned dictionaries and non-overlapping patches of size 8$\times$8 (CS to 12 samples). a) Original, b) Random (29.1 dbs), c) Unstructured/Structured (29.2 dbs), d) RIP-AB (32.1 dbs)}
\label{Figure_1}
\end{figure}

Table \ref{Table2} shows the mean peak signal to noise ratio (PSNR) of the reconstructed images (from the reconstructed {non-}overlapping patches), for three levels of signal compression, where
$
PSNR = 10 \, \log_{10} \big (  \frac{I^2_{max}}{MSE} \big ),
$
$I_{max}$  corresponding to the maximum image intensity. The RIP-AB sensing strategy achieves the best reconstruction PSNRs, as expected. Using a paired t-test, we found that the differences are statistically significant, with a p-value less than 0.001. Figure \ref{Figure_1} shows the original and reconstructed 8$\times$8 {non-}overlapping patches for a selected image (results with other selected images are also shown in the supplementary material, figures S1-S2). The quantitative improvements are visually observed as well. 
\subsection{Adaptive Statistical Compressive Sensing}
\label{subsec_expASCS}
We now compare the classification and reconstruction performance of the proposed two-step adaptive statistical CS framework (Section \ref{sec_adaptive}), with several configurations as indicated in Table \ref{Table1}. For this purpose, we generate two-class synthetic Gaussian signals of dimensions 36, 64, and 100, where the Bhattacharyya distance (BD) between them is varied as well, see Appendix \ref{App5} for details. 

In addition, we also use 6$\times$6, 8$\times$8, and 10$\times$10 non-overlapping patches extracted from 50 natural images from the Berkeley segmentation dataset \cite{BerkelyDataSet}. Three levels of noise were also considered: no added noise, noise level of 40 dbs, and noise level of 30 dbs, for both the synthetic and natural image signals. There are 19 learned classes in this case, used to efficiently model patches of natural images (see \cite{Guoshen2010} for details). 

The dictionary is here not adapted to each image, in order to provide the same GMM to all the considered two-step configurations (Table \ref{Table1}), so that the differences between configurations are due only to the sensing itself and not due to the adapted dictionaries. 
\subsubsection{Synthetic Signals}

\begin{figure}[!hbp]
\begin{center}
\scriptsize
\includegraphics[width=0.5\textwidth]{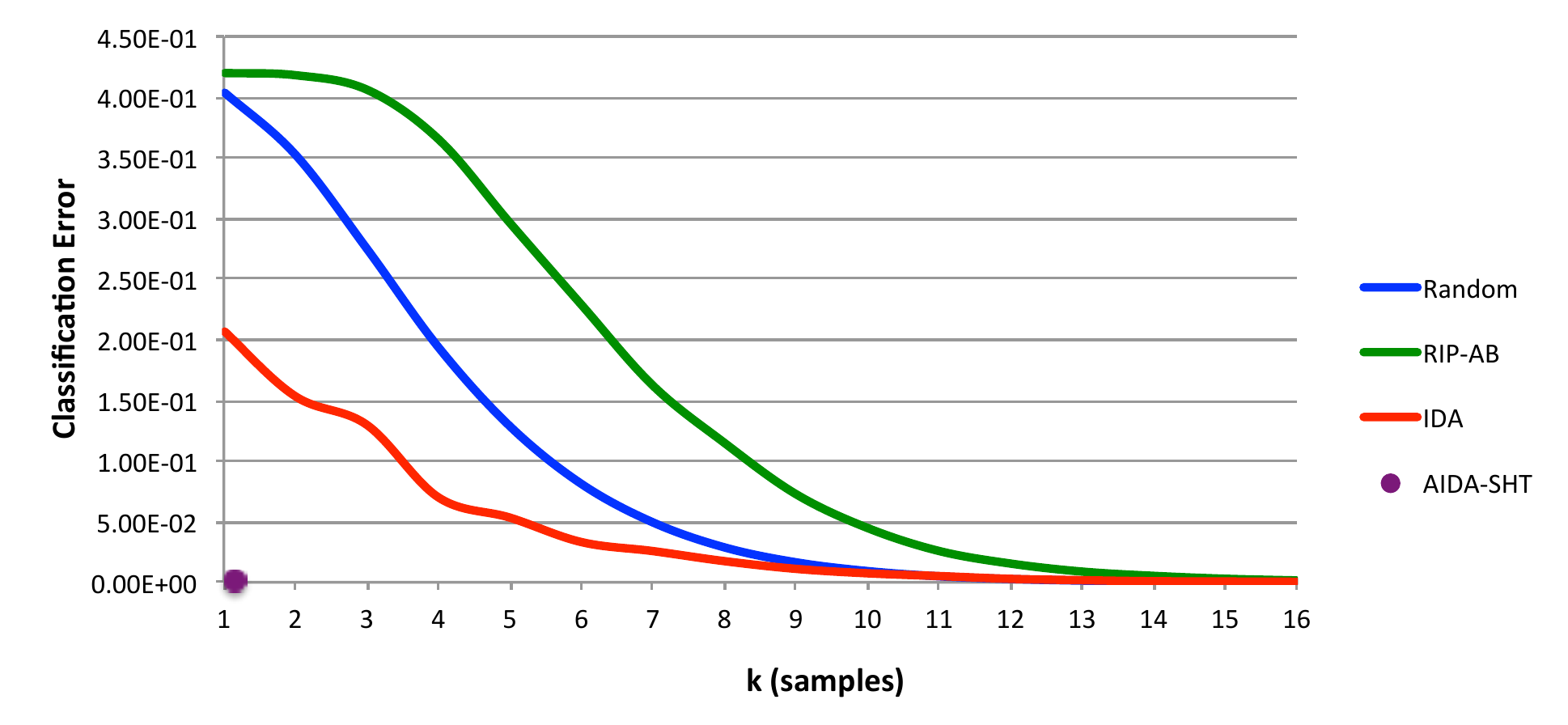}%
\includegraphics[width=0.5\textwidth]{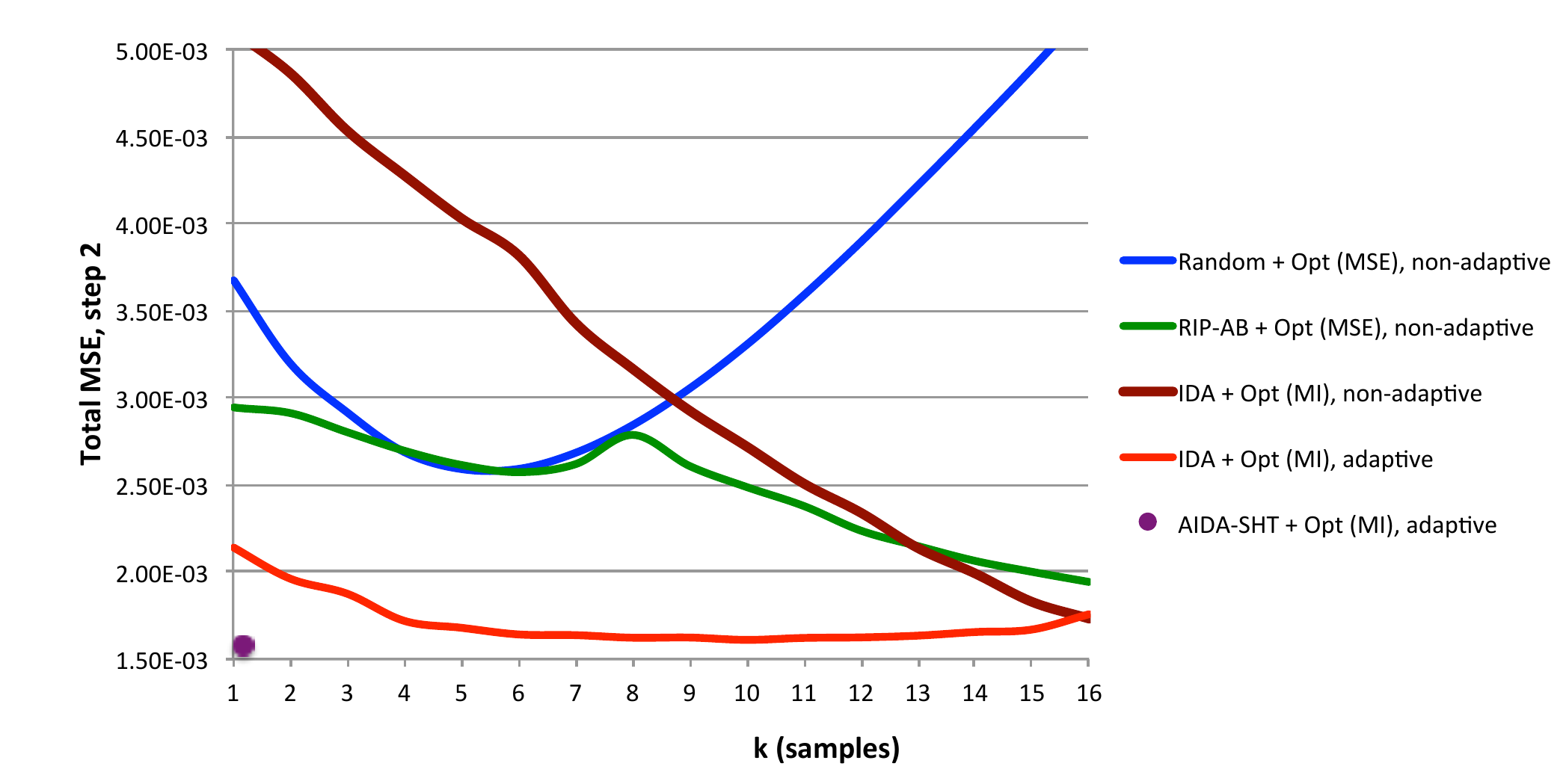}\\
(a)\\
\includegraphics[width=0.5\textwidth]{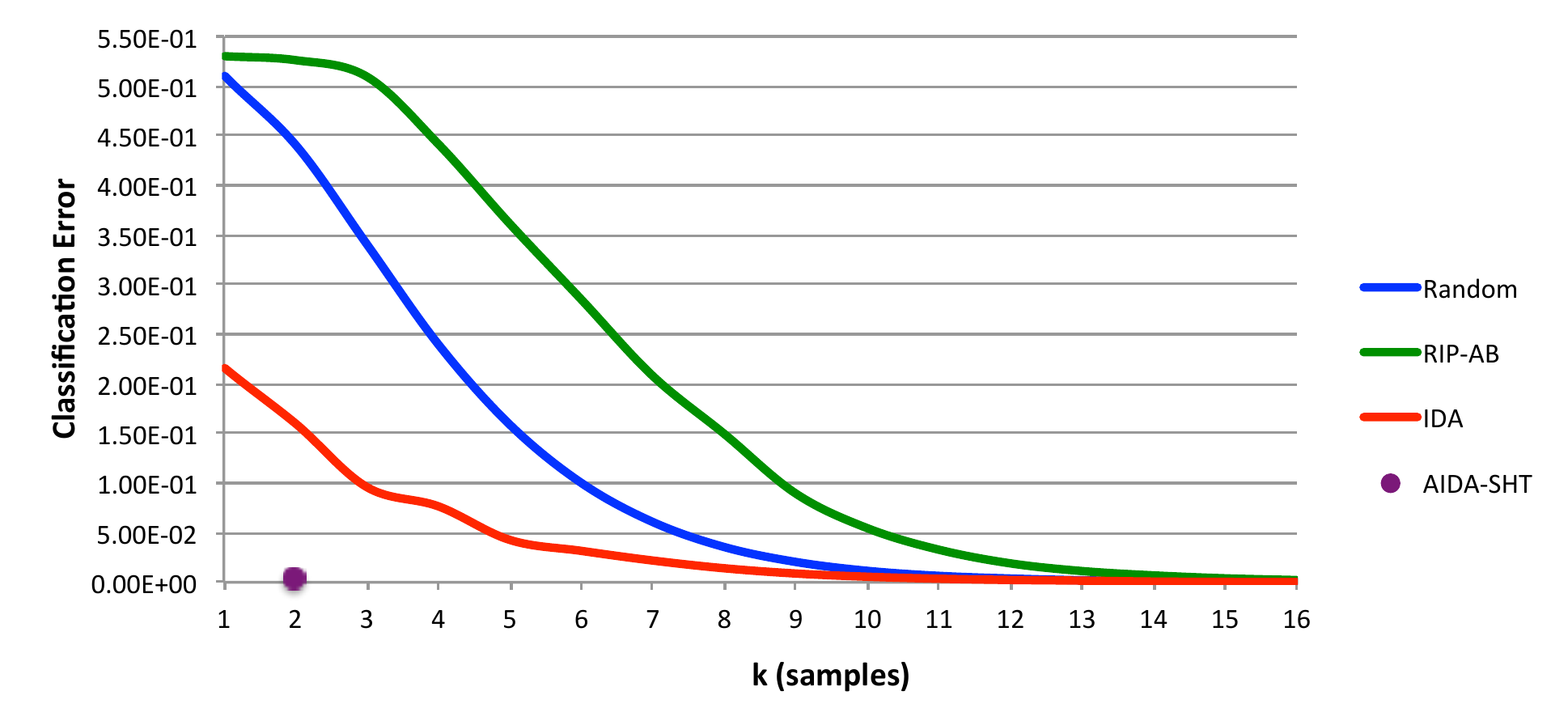}%
\includegraphics[width=0.5\textwidth]{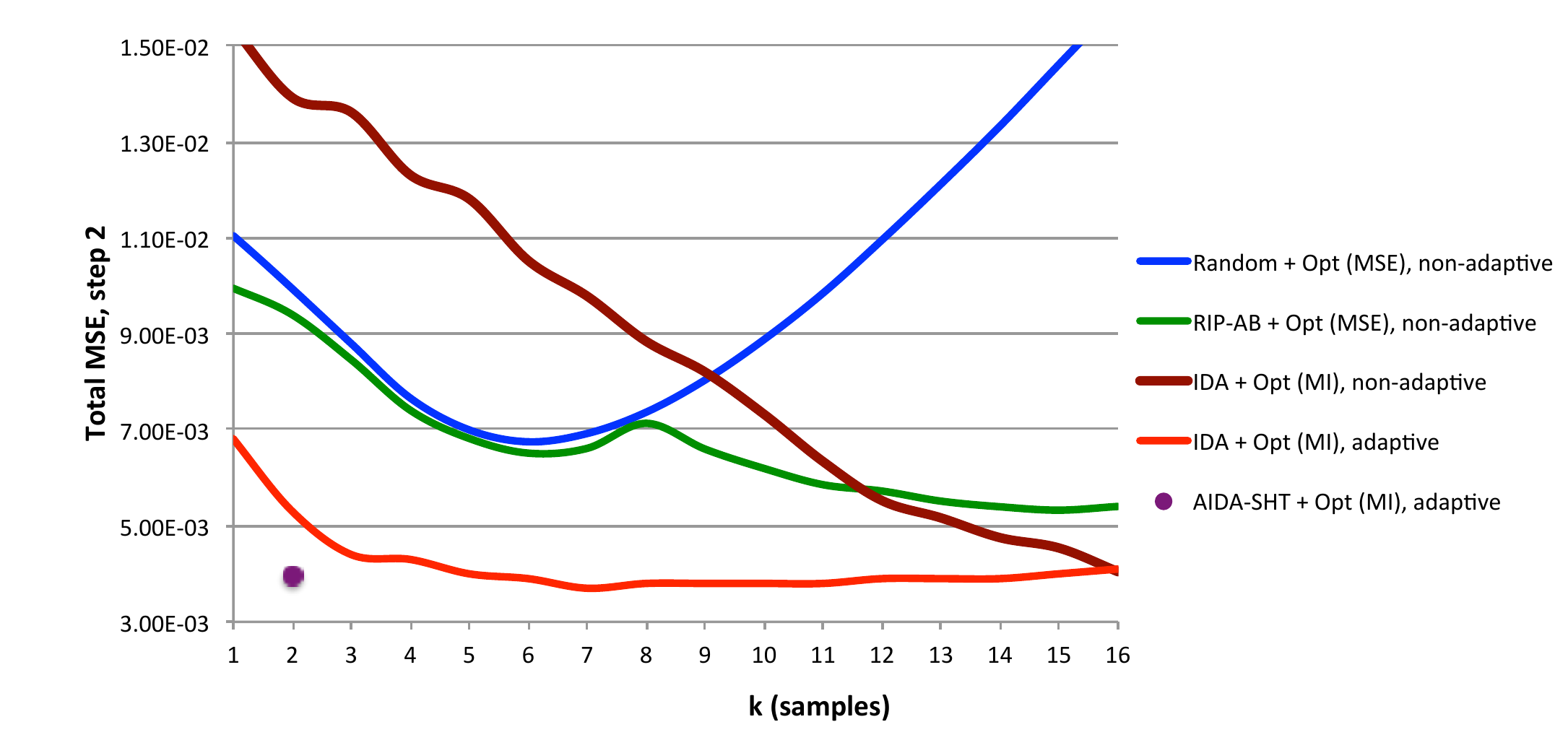}
(b) \\
\includegraphics[width=0.5\textwidth]{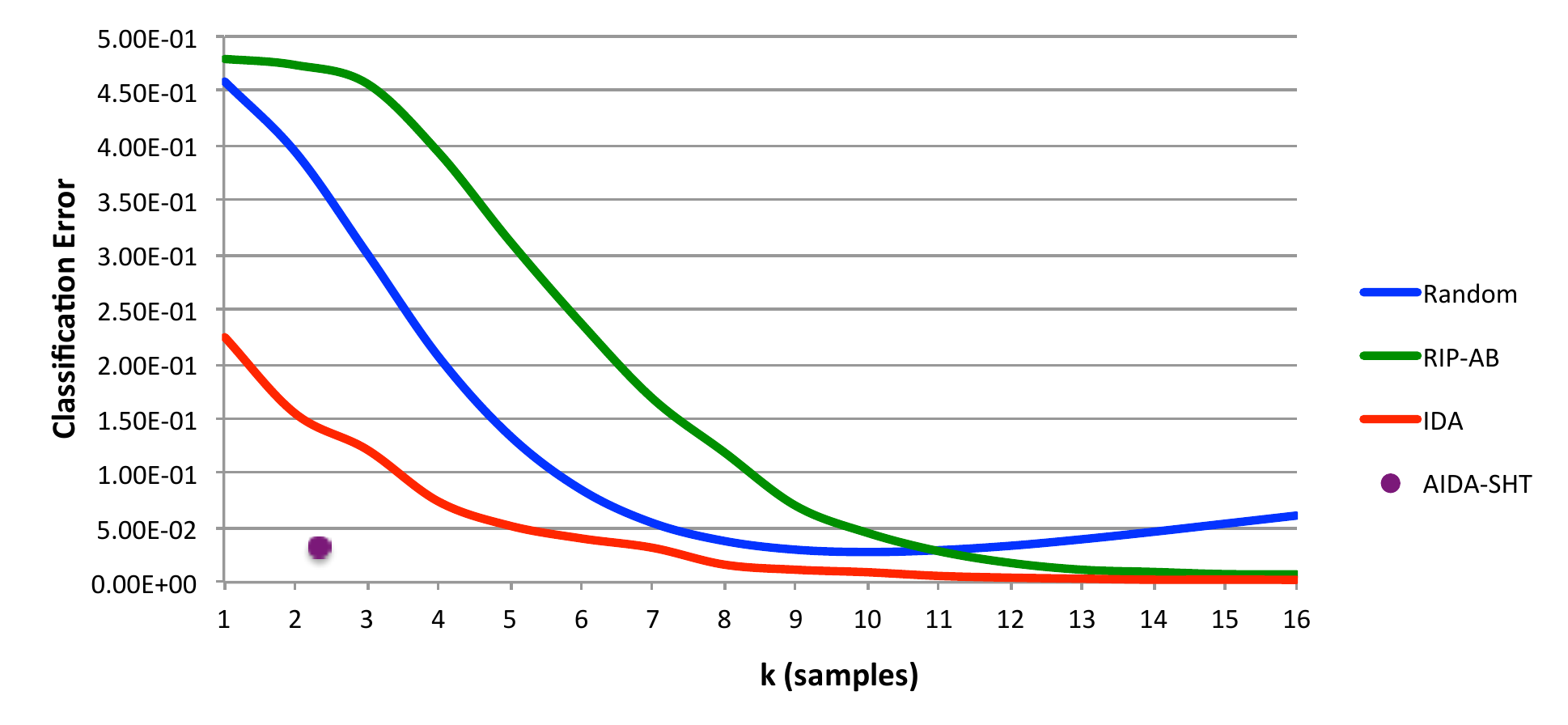}%
\includegraphics[width=0.5\textwidth]{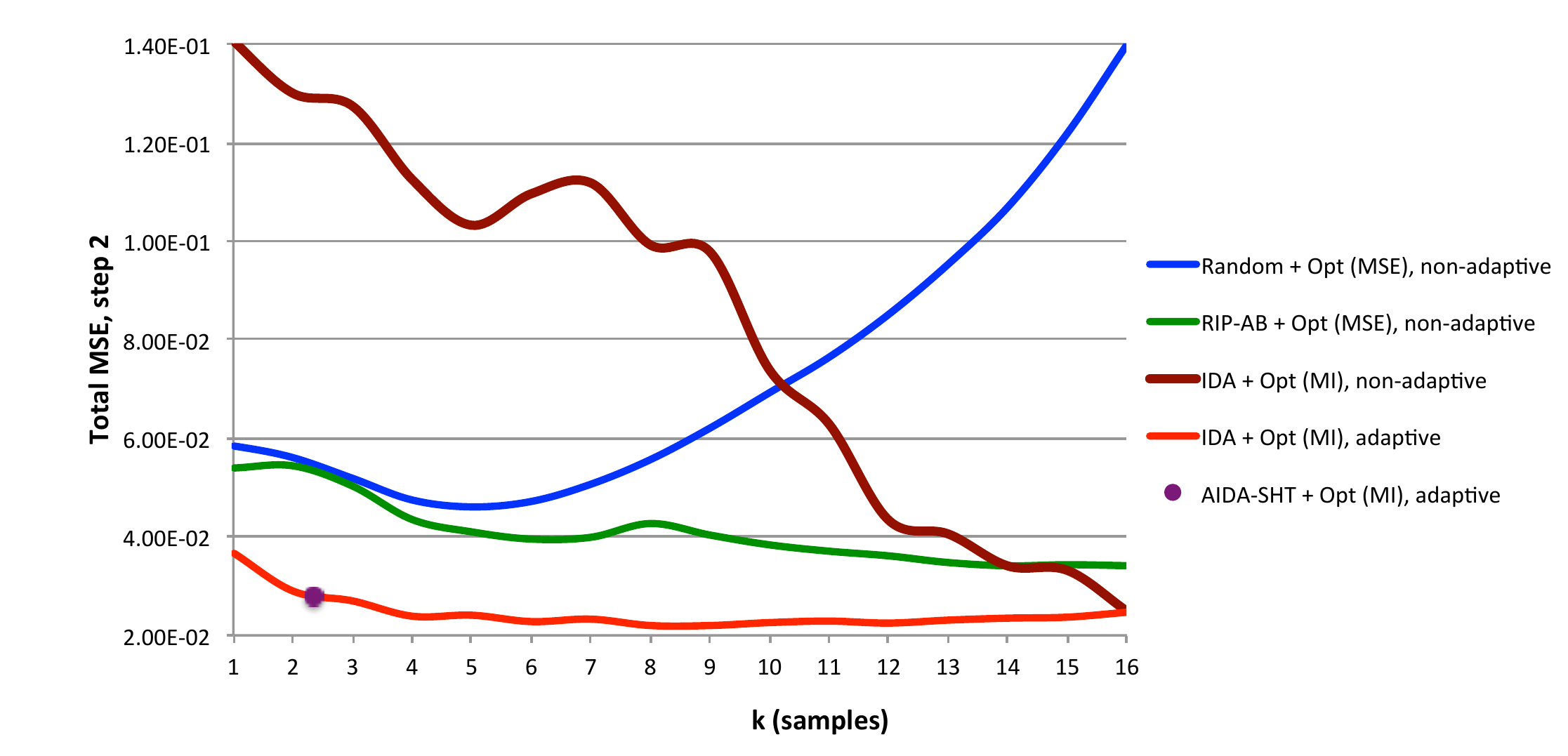}
(c) \\
\end{center}
\caption{Classification accuracy (Step 1) and reconstruction MSE (step 2) for synthetic signals of dimension 64 (CS to 16 samples) and BDs $\in [30 ~ 46)$. a) No noise, b) SNR of 40 dbs, c) SNR of 30 dbs.}
\label{Figure_2}
\end{figure}

Figure \ref{Figure_2} shows the classification error at Step 1 and the corresponding MSE at Step 2, for the indicated BDs and three levels of noise. As expected, the best classification results are obtained for IDA and AIDA-SHT, being AIDA-SHT the best, stopping automatically between two and three samples, on average. The worst classification results were obtained for the RIP-AB non-adaptive sensing matrix and random sensing. It can be noticed that despite the relatively bad classification performance of RIP-AB, the reconstruction is equal or better than random and it improves as $K \to M$, as expected, since RIP-AB is better than random for non-adaptive single-step batch CS (sections \ref{sec_nonAdaptive} and \ref{subsec_expBlockSCS}), i.e., when $K = M$. RIP-AB does better than random on the final reconstruction, using $K < M$ measurements, because the BDs between the two Gaussians is relatively small, hence, selecting the wrong Gaussian will not be that harmful for reconstruction during the second step. The lowest reconstruction errors are achieved using AIDA-SHT in the first step and the optimal (in the MI sense) adaptive sensing in the second step. However, as the noise increases, the reconstruction performance of AIDA-SHT degrades with respect to IDA. The reason for this is that IDA uses only the (clean) information provided by the covariance matrices, while SHT tests hypotheses based on the noisy CS signals. Nevertheless, AIDA-SHT stops automatically, without knowing a priori the number of steps $K$ required in the first step, while IDA (or AIDA alone) requires a pre-defined number of steps $K$. 

It can also be observed in Figure \ref{Figure_2} that using IDA in the first step and the non-adaptive optimal for the selected Gaussian in the second step is worse than using random in the first step and the non-adaptive optimal for the selected Gaussian in the second, despite the fact that IDA has a very good classification performance in the first step. The reason for this is that IDA uses information from the average covariance matrix, which is likely to lead to a sensing matrix in the first step that is correlated with the sensing matrix in the second, especially when the number of classes is small (two in this case). Hence, the importance of adaptivity in the second step. This does not affect random sensing, since it is very unlikely that a random sensing matrix will be correlated with the non-adaptive optimal sensing matrix in the second step. This does not seem to affect the RIP-AB sensing, probably because it does not use the average covariance matrix either. 

\begin{figure}[!hbp]
\begin{center}
\scriptsize
\includegraphics[width=0.5\textwidth]{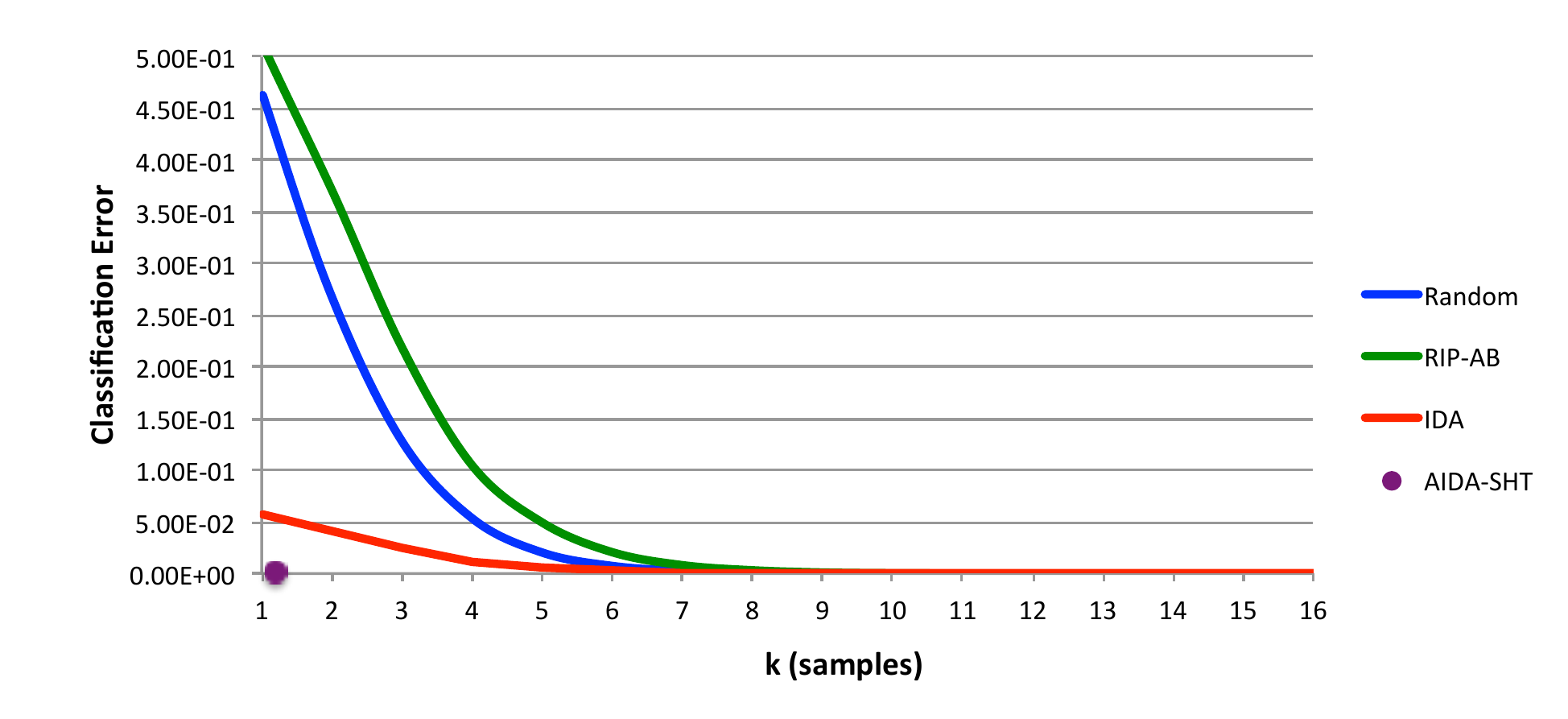}%
\includegraphics[width=0.5\textwidth]{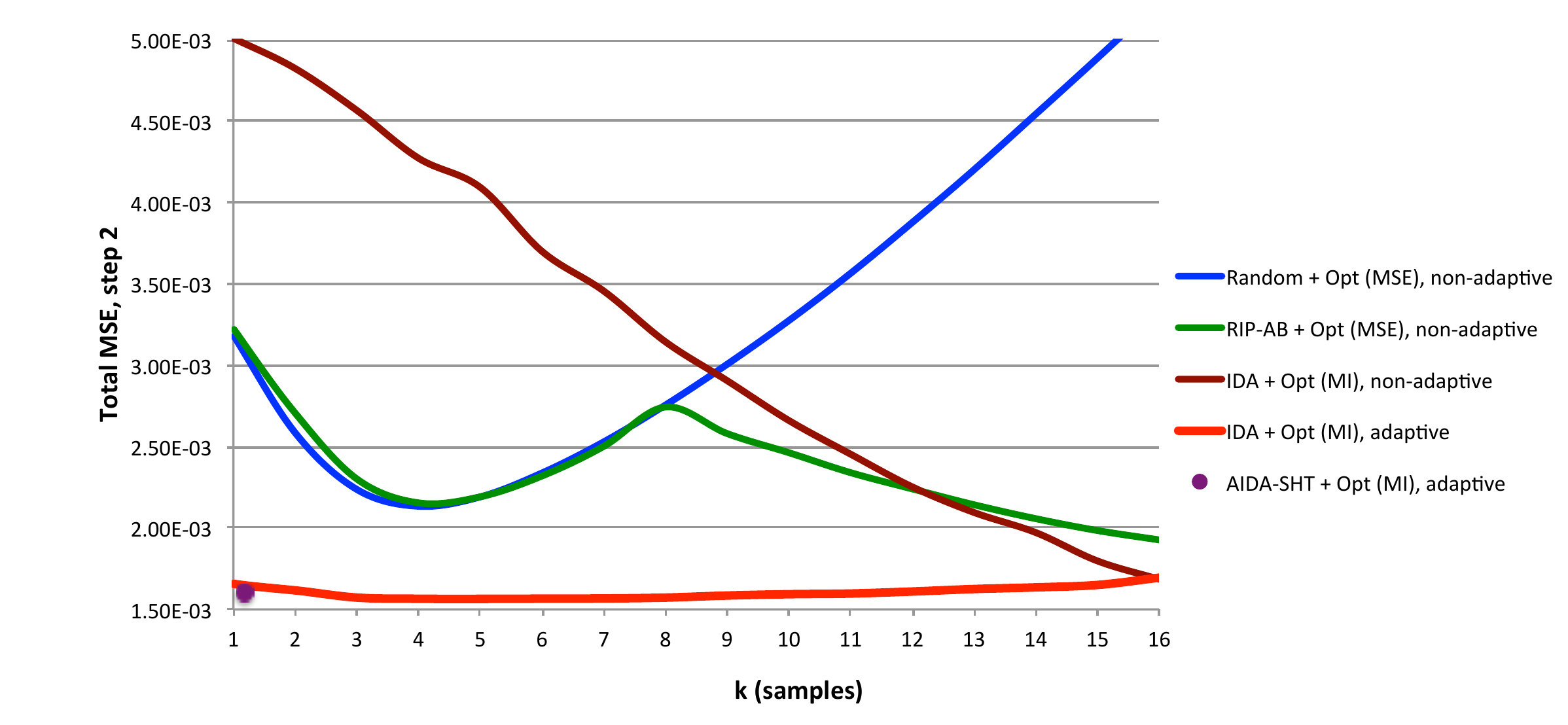}
(a) \\
\includegraphics[width=0.5\textwidth]{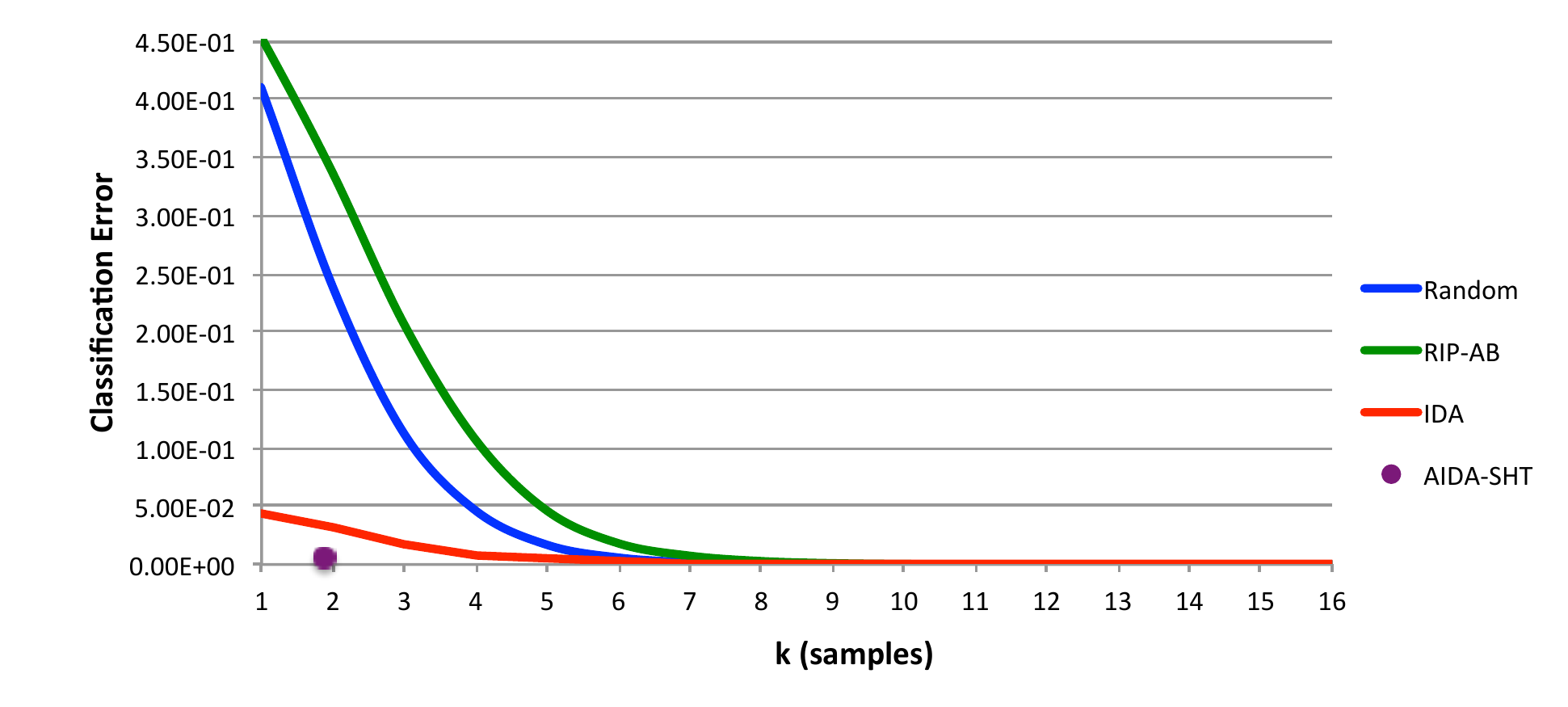}%
\includegraphics[width=0.5\textwidth]{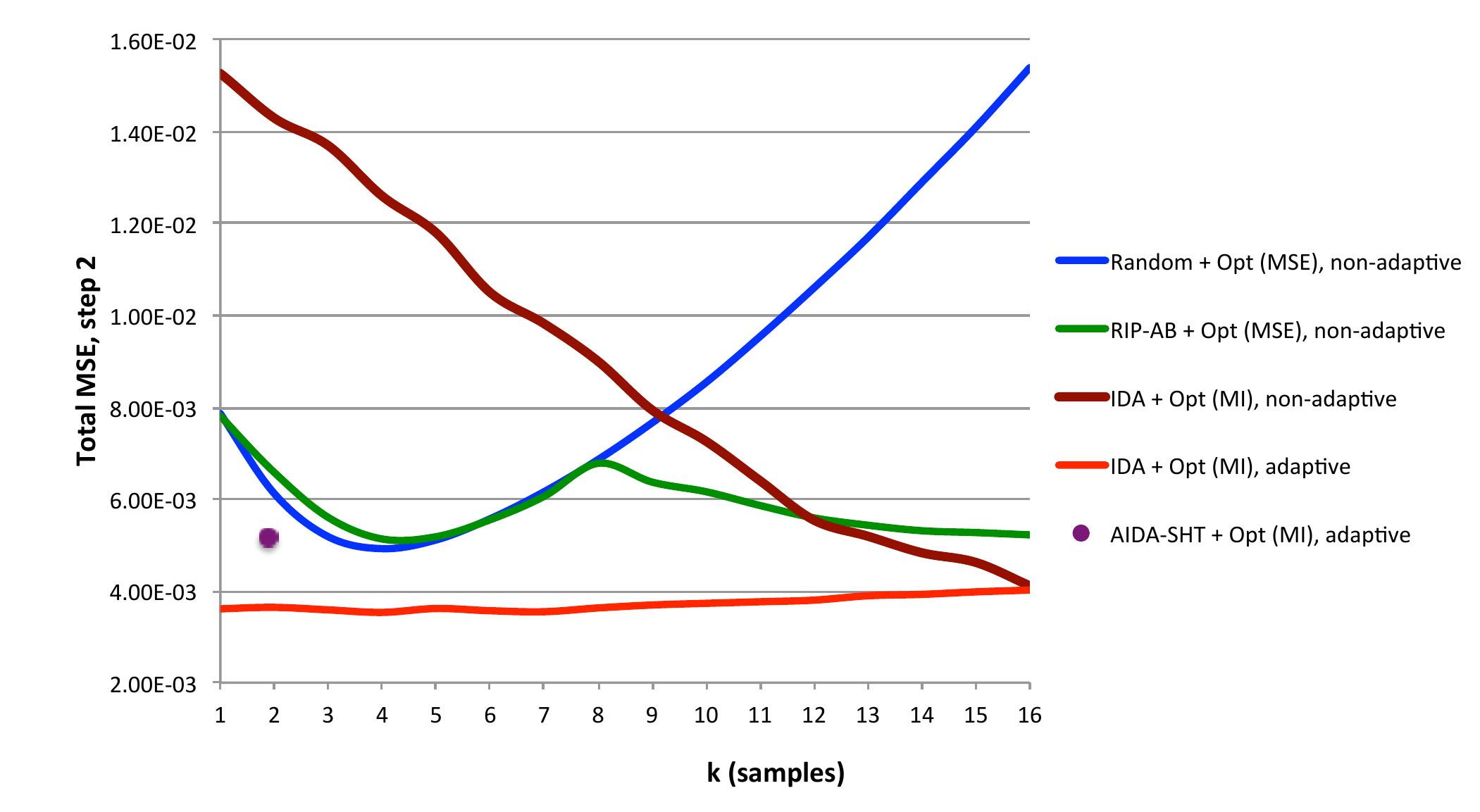}
(c) \\
\includegraphics[width=0.5\textwidth]{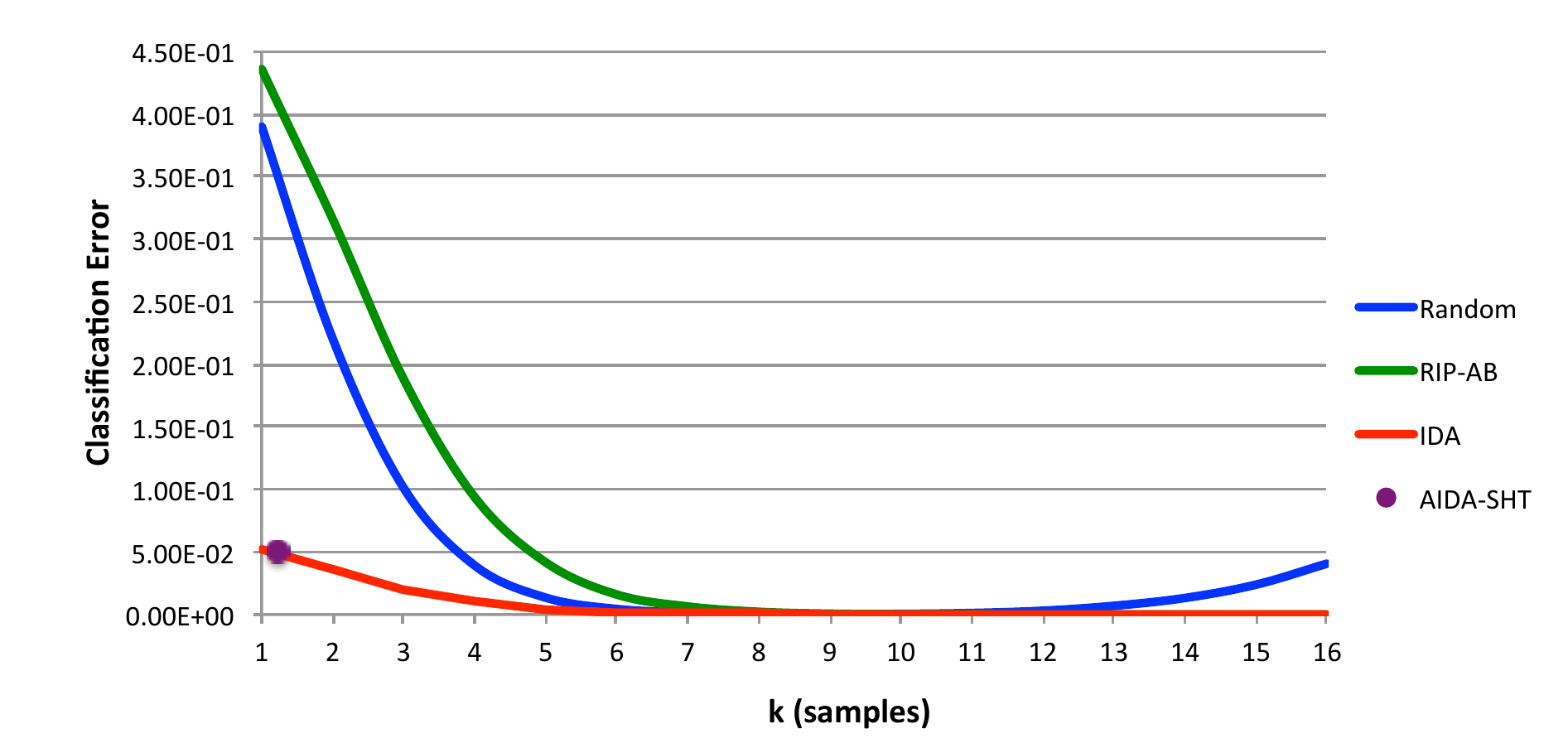}%
\includegraphics[width=0.5\textwidth]{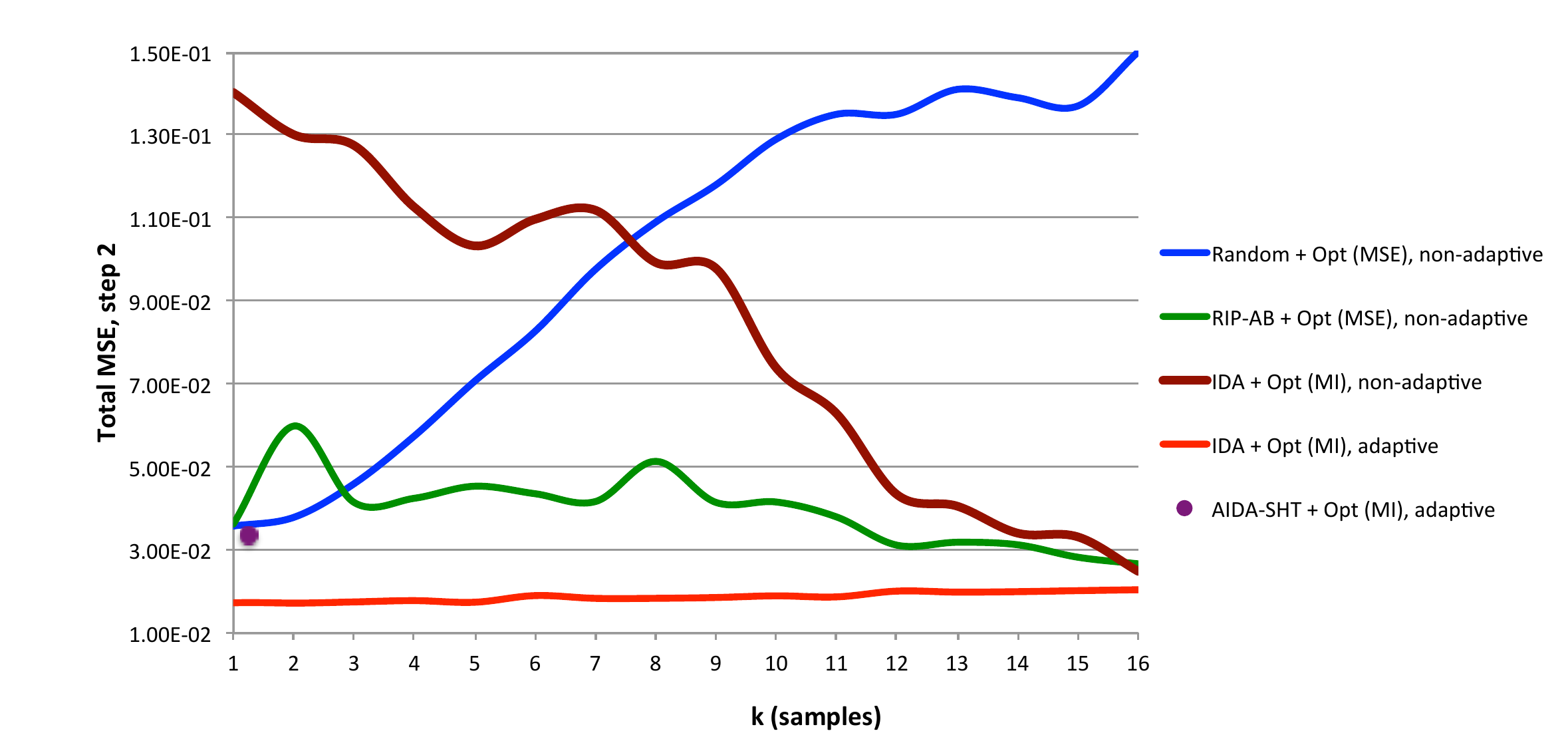}
(c)\\
\end{center}
\caption{Classification accuracy (Step 1) and reconstruction MSE (step 2) for synthetic signals of dimension 64 (CS to 16 samples) and BDs $\in [62 ~ 78)$. a) No noise, b) SNR of 40 dbs, c) SNR of 30 dbs.}
\label{Figure_3}
\end{figure}

Figure \ref{Figure_3} shows the classification error at the first step and the corresponding reconstruction MSE at the second step, for larger BDs than those in Figure \ref{Figure_2}, with the same noise levels. All classification accuracies are now better than in Figure \ref{Figure_2}, since now the distance between the covariance matrices is larger, making classification easier. AIDA-SHT still has the best classification accuracy, but the difference with IDA is reduced. In the second step, the reconstruction performance using AIDA-SHT in the first step deteriorates more with noise, and in this case IDA is better. We believe the reason for this is that the adaptive sensing matrix of AIDA is much better than IDA for classification, but good features for classification are not necessarily good for reconstruction. In addition, since at larger BD distances classification becomes easier, the need for a good classifier in the first step reduces, while reconstruction becomes more important. 

Due to space limitations, the results for other signal sizes and BDs are only shown in the supplementary material. However, and as can be seen from the figures in the supplementary material (figures S3-S40), the behavior is quite similar to the results just presented. 

\subsubsection{Patches from Natural Images}

\begin{figure}[!hbp]
\begin{center}
\scriptsize
\includegraphics[width=0.8\textwidth]{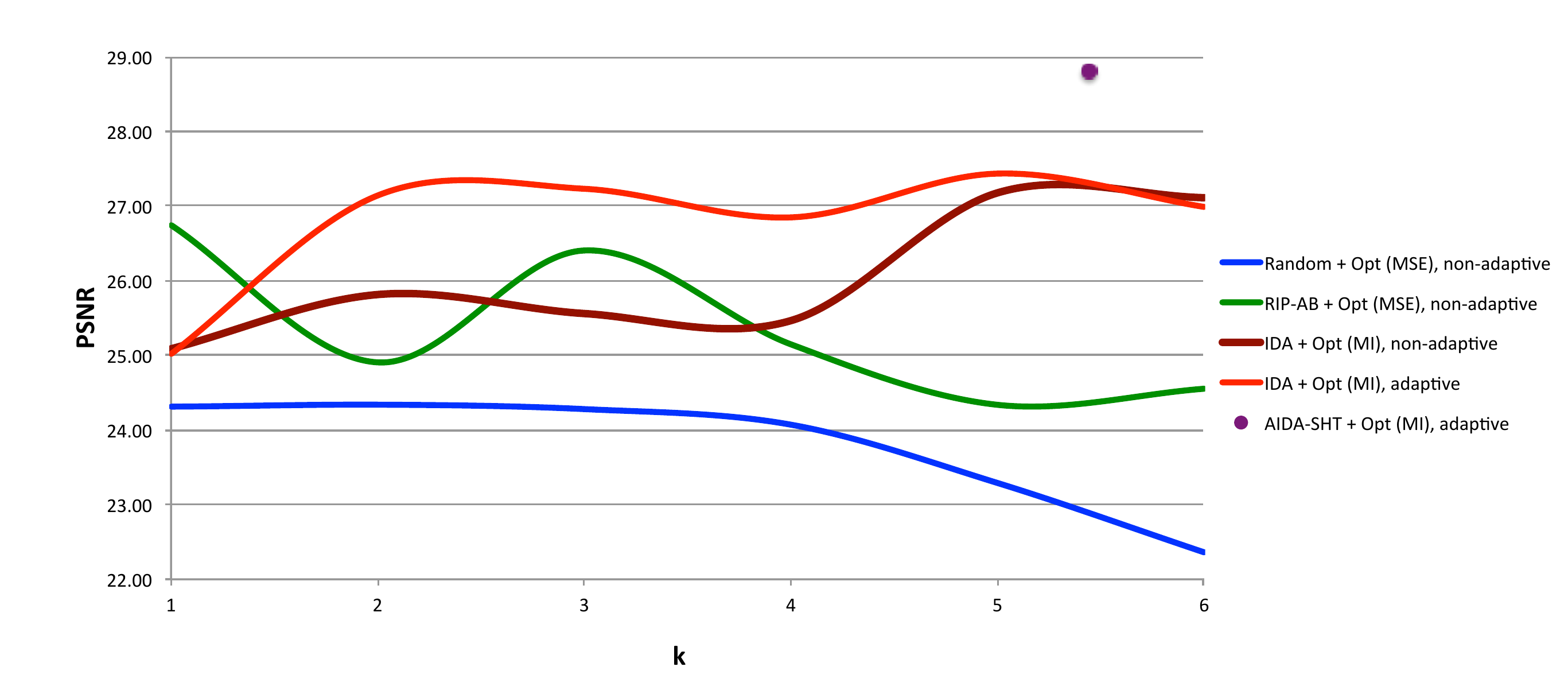}\\
(a)\\
\includegraphics[width=0.8\textwidth]{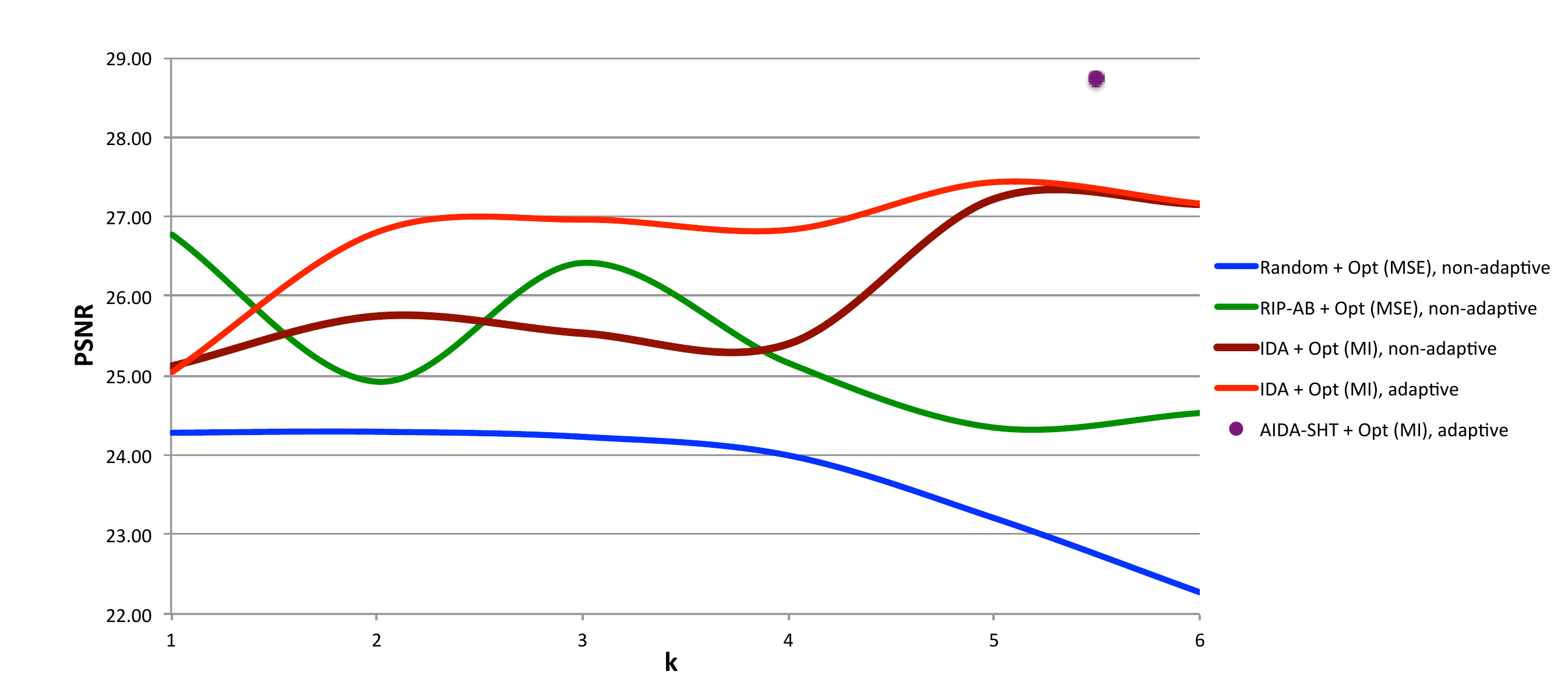}\\
(b)\\
\includegraphics[width=0.8\textwidth]{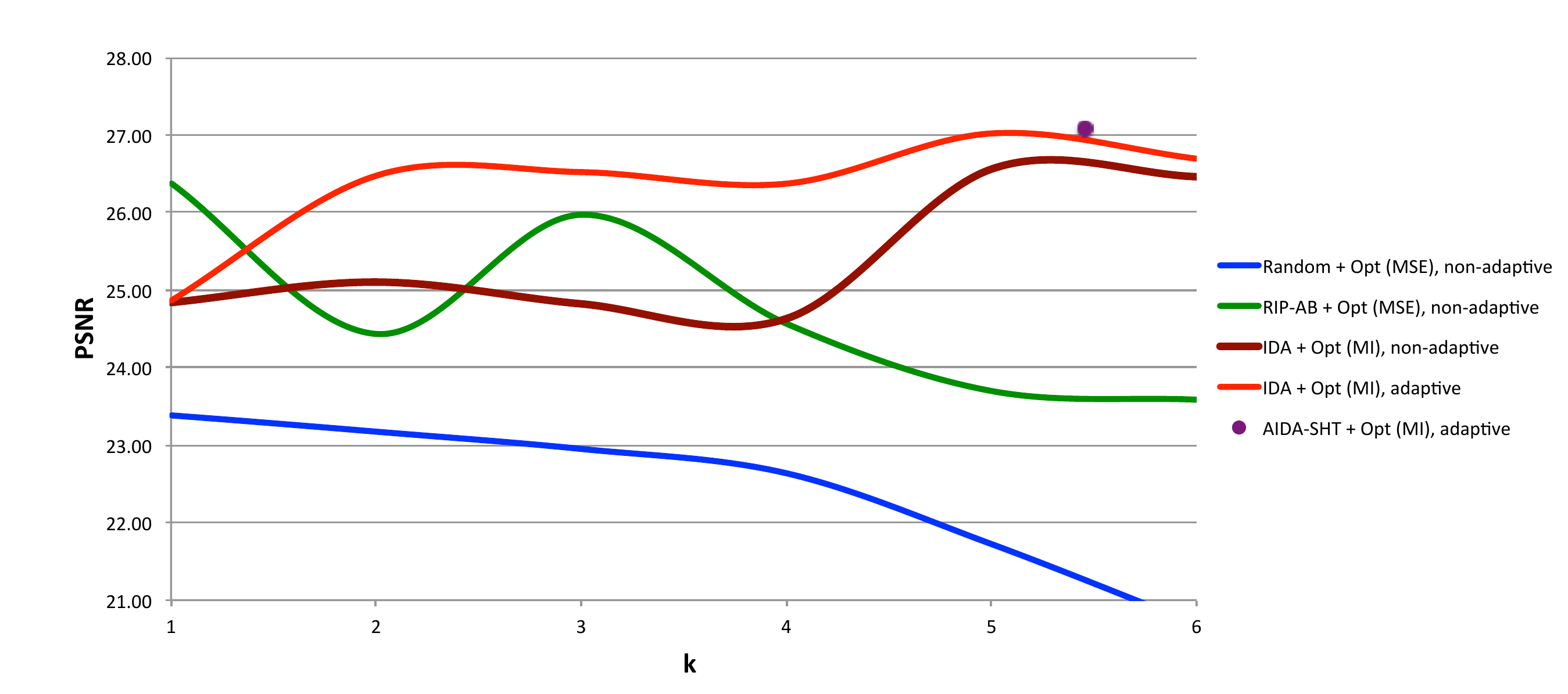}\\
(c)
\end{center}
\caption{PSNR (Step 2) of reconstructed natural images with non-overlapping patches of size 6$\times$6 (CS to 6 samples). a) No noise, b) SNR of 40 dbs, c) SNR of 30 dbs.}
\label{Figure_4}
\end{figure}

Figure \ref{Figure_4} shows the mean PSNR for the non-overlapping $6 \times 6$ patches extracted from 50 natural images at the three noise levels considered. Here we do not know the right class for each one of the signals, hence, we only report the reconstruction accuracy in the second step. The best reconstruction performance in the noise-free case is obtained for AIDA-SHT, and the classification (first step) stops between 5 to 6 samples on average. As the noise increases, the difference between AIDA and IDA reduces. This follows the same behavior observed with synthetic data, where AIDA-SHT degrades with noise. There is a wave-like behavior on the non-adaptive protocols used, with the exception of random sensing, probably due to the possible coherence of the sensing matrix between steps, which depends on $K$. Results for non-overlapping patches of sizes 8$\times$8 and 10$\times$10 can be found in the supplementary material (figures S41 and S42).     

\begin{figure}[!hbp]
\begin{center}
\scriptsize
\includegraphics[width=0.48\textwidth, height=0.37\textwidth]{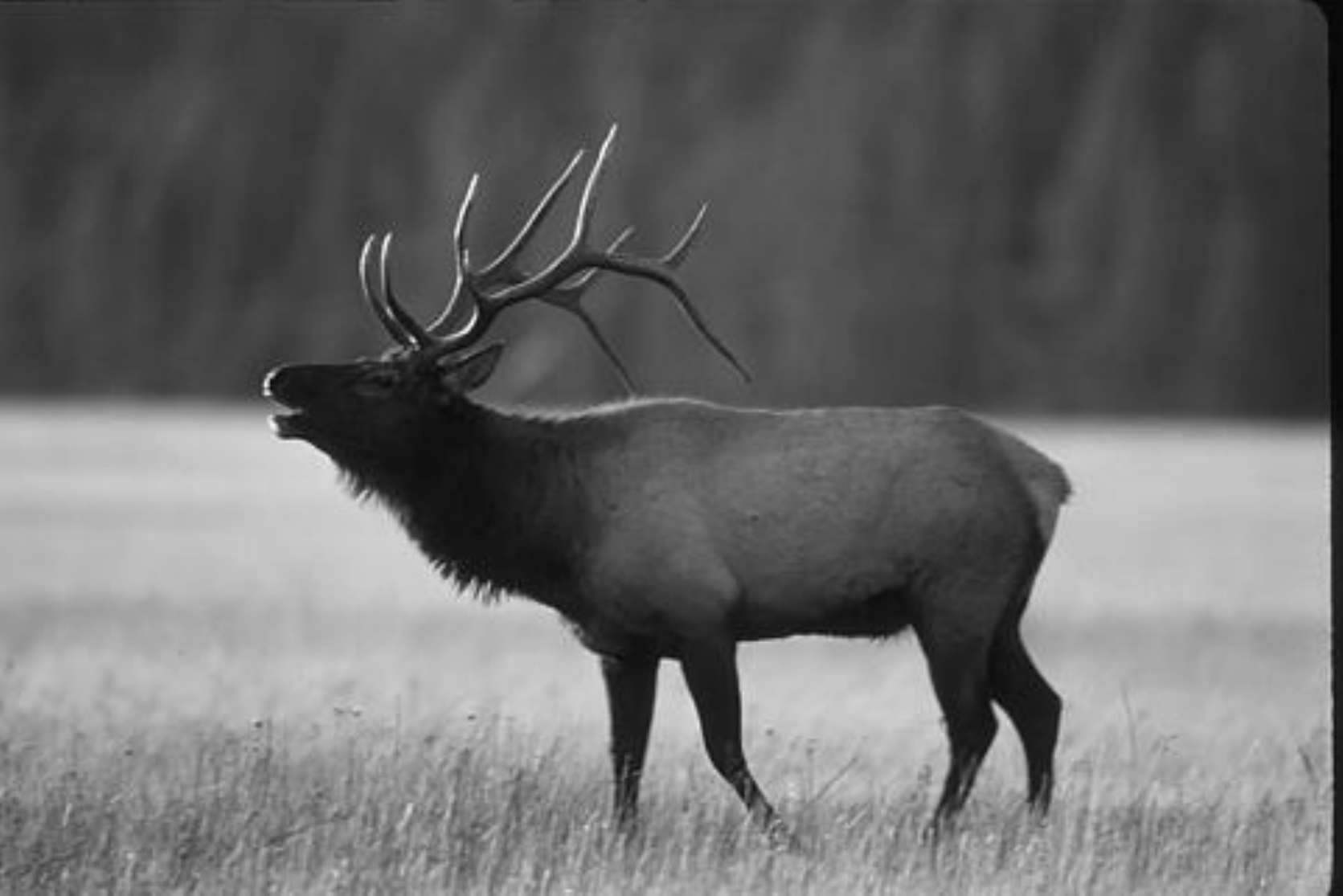}%
\includegraphics[width=0.48\textwidth, height=0.37\textwidth]{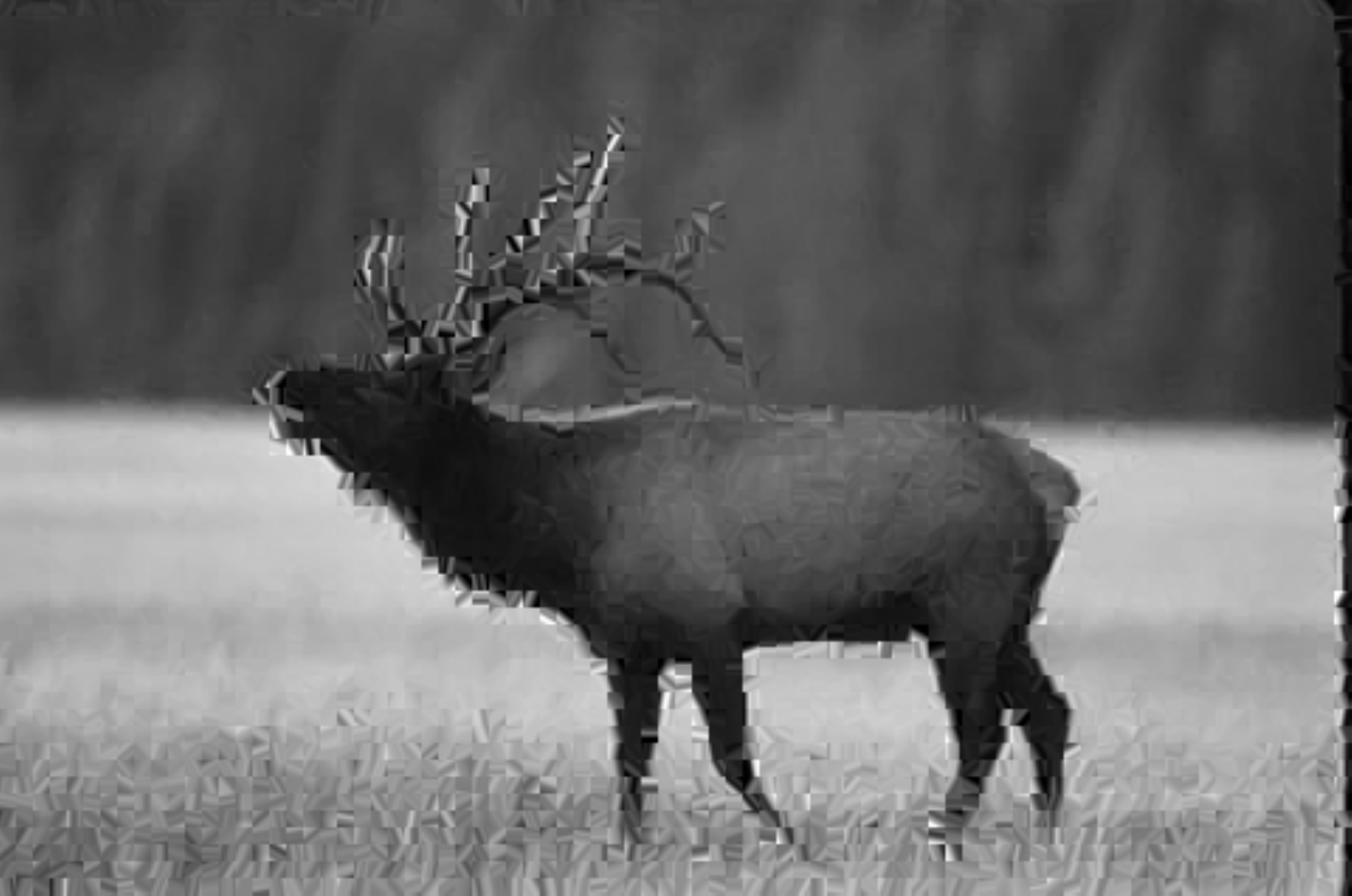}\\
(a) \qquad \qquad \qquad \qquad \qquad \qquad \qquad \qquad \qquad \qquad \qquad \qquad \qquad \qquad (b)\\
\includegraphics[width=0.48\textwidth, height=0.37\textwidth]{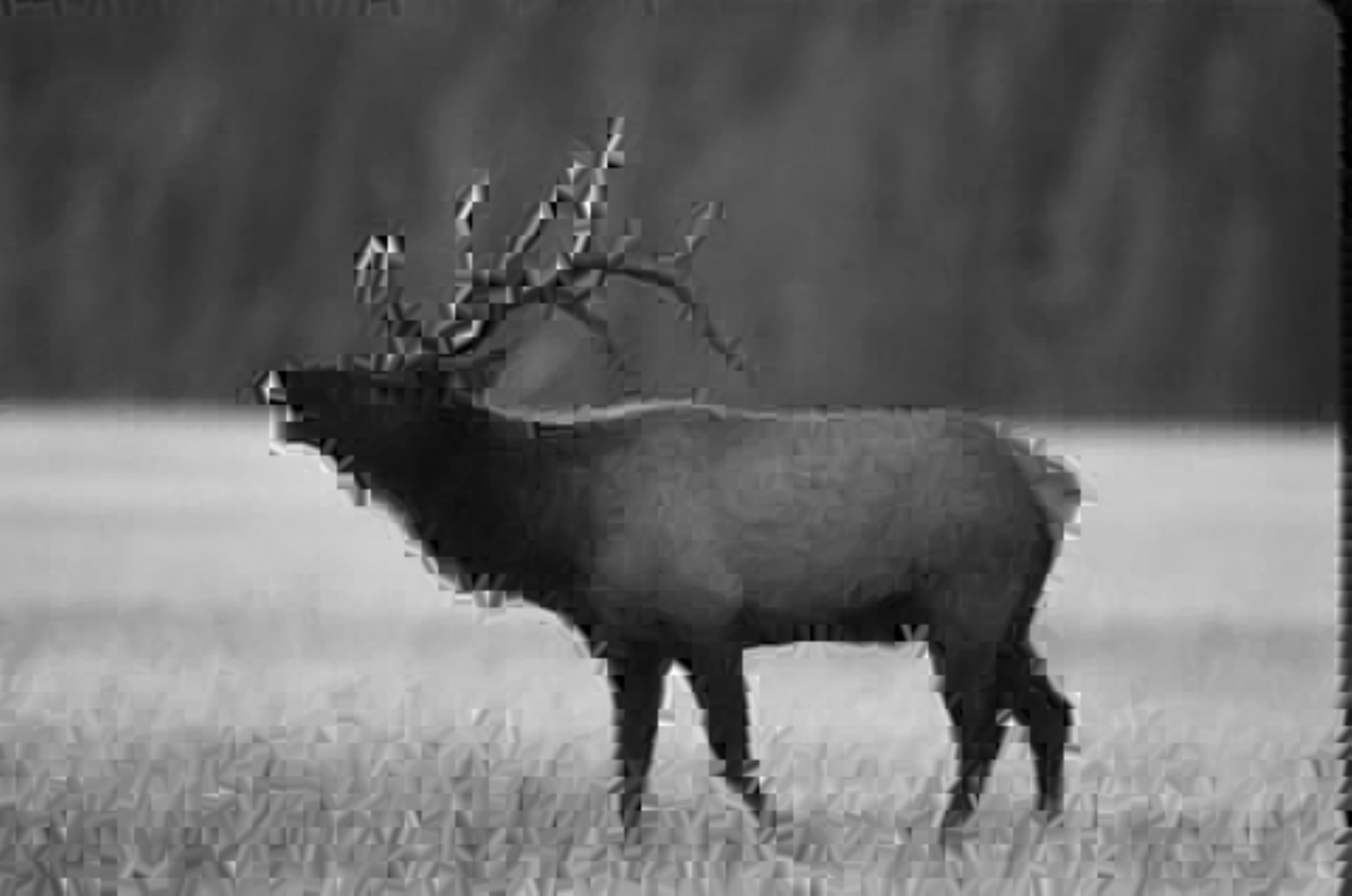}%
\includegraphics[width=0.48\textwidth, height=0.37\textwidth]{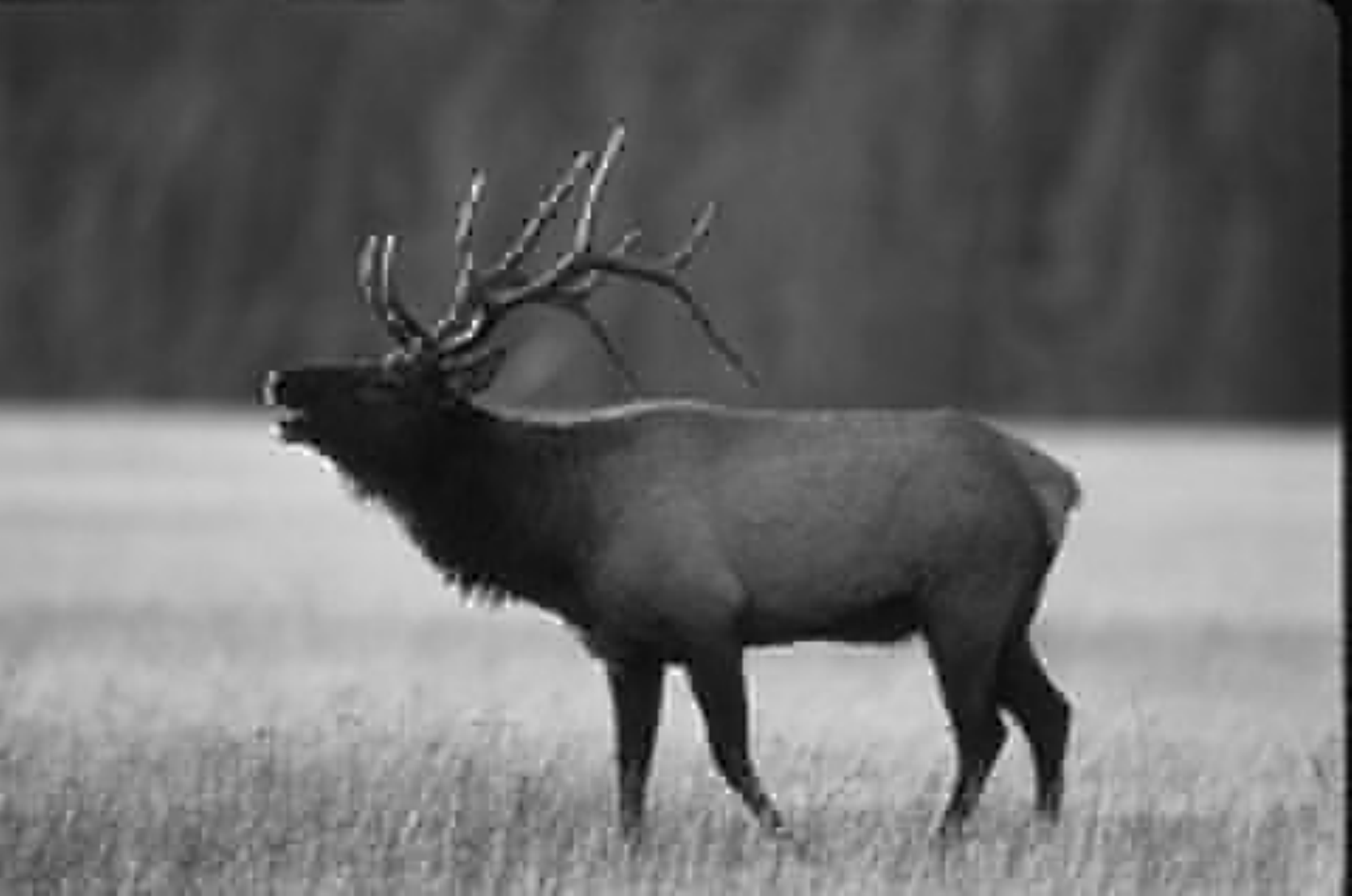}\\
(c) \qquad \qquad \qquad \qquad \qquad \qquad \qquad \qquad \qquad \qquad \qquad \qquad \qquad \qquad (d)\\
\includegraphics[width=0.48\textwidth, height=0.37\textwidth]{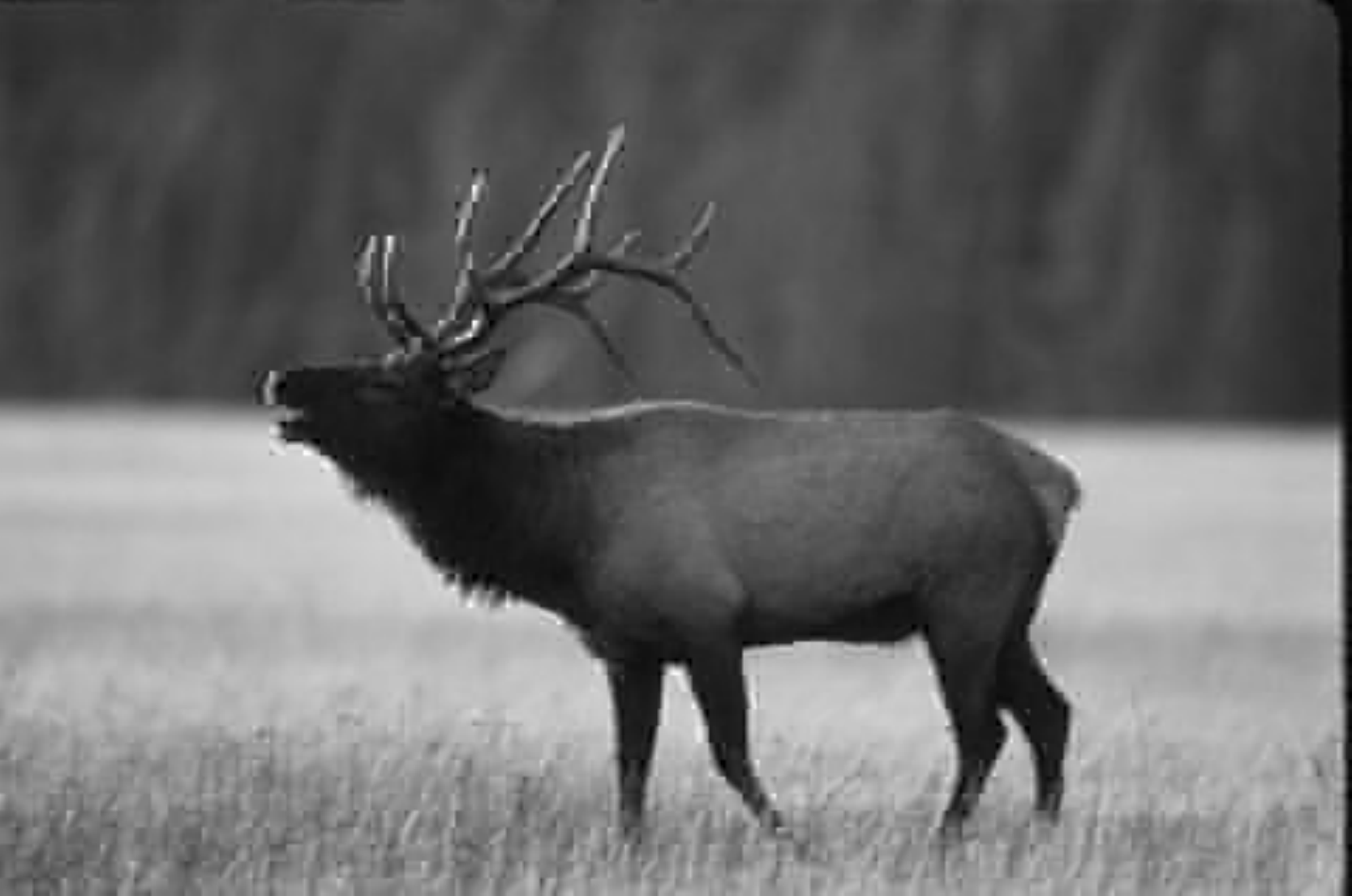}%
\includegraphics[width=0.48\textwidth, height=0.37\textwidth]{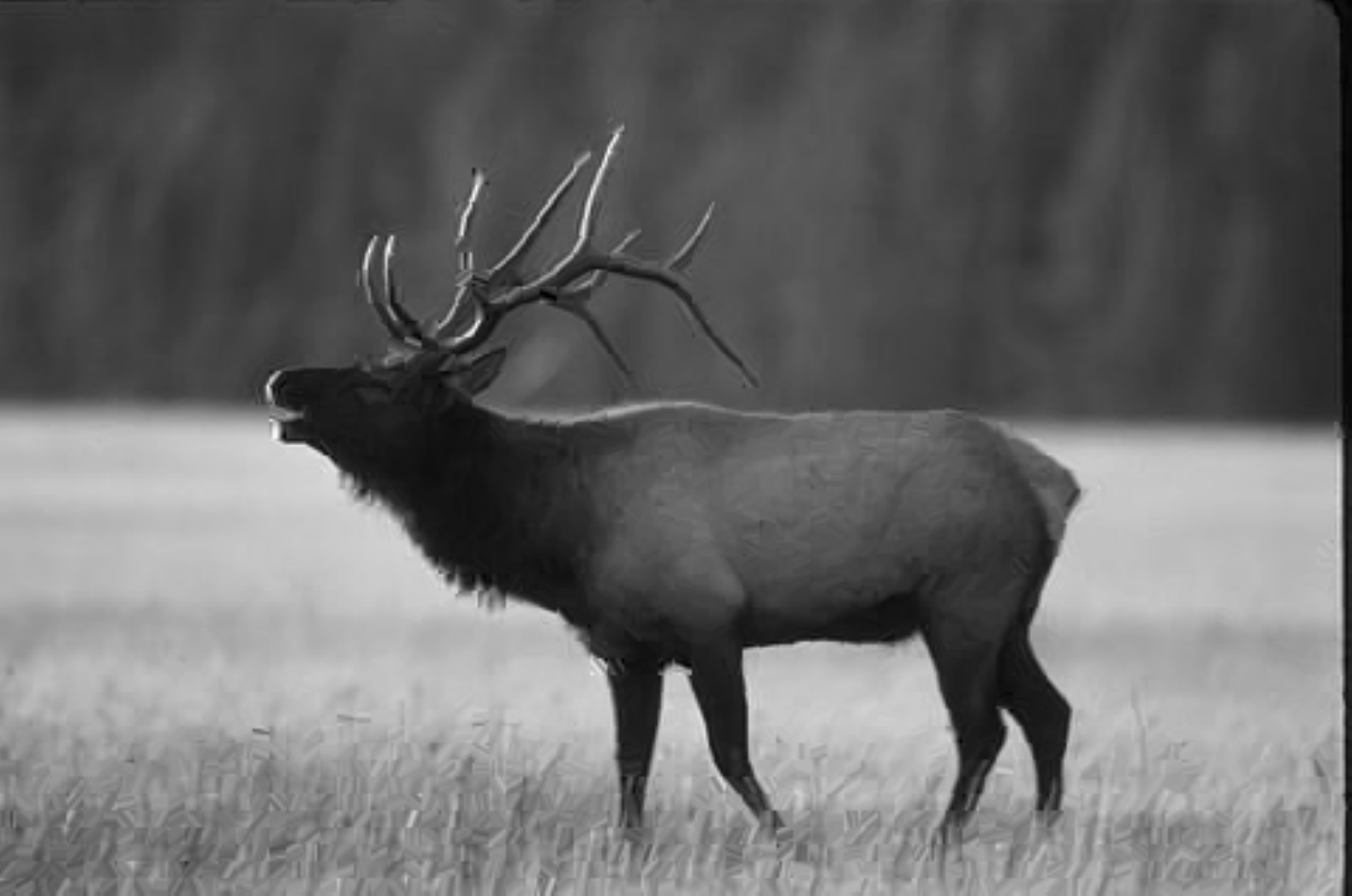}\\
(e) \qquad \qquad \qquad \qquad \qquad \qquad \qquad \qquad \qquad \qquad \qquad \qquad \qquad \qquad (f)
\end{center}
\caption{Reconstructed image from non-overlapping patches of size 6$\times$6 (CS to 6 samples) using the following two-step protocols: a) Original, b) Random - Optimum (MSE) non-adaptive (26.8 dbs), c) RIP-AB - Optimum (MSE) non-adaptive (27.66 dbs), d) IDA - Optimum (MSE) non-adaptive (30.51 dbs), e) IDA- Optimum (MI) adaptive (30.77 dbs), and f) AIDA-SHT-Optimum (MI) adaptive (33.9 dbs).}
\label{Figure_5}
\end{figure}

Figure \ref{Figure_5} shows the original and reconstructed $6 \times 6$ non-overlapping patches for a selected image using the different two-step CS configurations considered (Table \ref{Table1}), $K=5$, and no noise added (AIDA-SHT uses $K=5.41$ on average). Using IDA and AIDA-SHT on the first step and the optimal adaptive (MI) on the second step produces the best reconstruction performances, with PSNRs that are $\sim 6$ dbs above random sensing in the first step. Notice that the reported reconstructions do not use dictionary (and sensing) adaptation, explaining the relative bad quality of the reconstructed patches using random and RIP-AB sensing in the first step (compare to the results reported on Section \ref{subsec_expBlockSCS}). Results with other selected images and patch sizes can be found in the supplementary materials (figures S43-S53).

{Note that the results using the proposed two-step framework (figures \ref{Figure_2}-\ref{Figure_4} and S3-S42) also include the single-step batch results for random, RIP-AB, and IDA, for the particular case when $K=M$. Hence, from these figures, we can conclude that in general the proposed two-step framework does provide better reconstructions than single-step (batch) random, RIP-AB, and IDA. AIDA-SHT should not be done in batch mode, since it is fully adaptive.}  

In summary, AIDA-SHT in the first step and the adaptive optimal (MI) sensing matrix in the second is the best sensing protocol for the two-step SCS proposed, provided that the noise level is not too high. A good alternative (less expensive computationally) to AIDA-SHT in the first step is the well-known non-adaptive batch IDA, provided that we know a priori the optimal number of measurements $K$ needed for class detection in the first step. 

\subsection{Classification}
\label{subsec_expClass}
\begin{figure}[!hbp]
\begin{center}
\scriptsize
\includegraphics[width=0.8\textwidth]{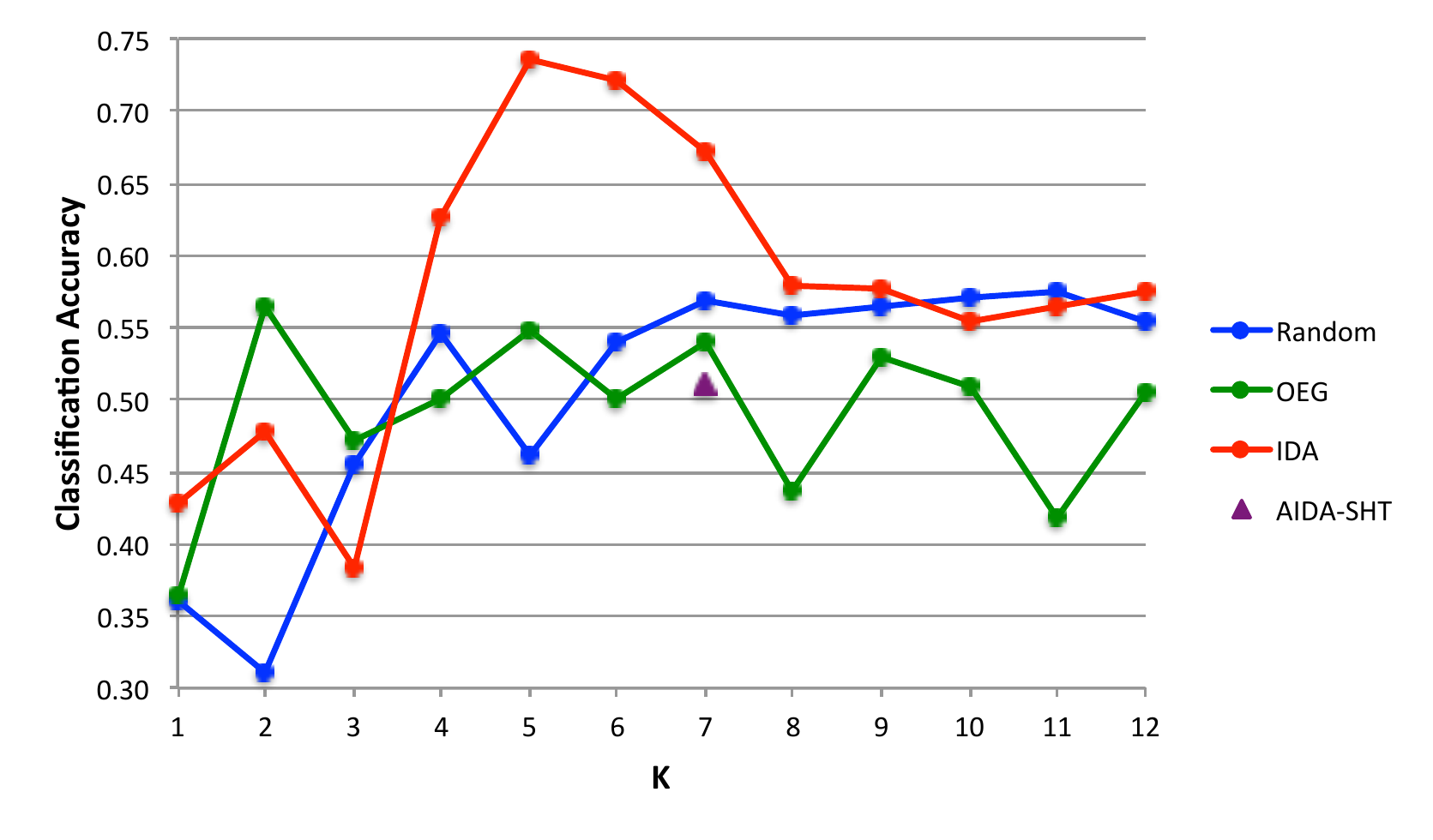}\\
(a) \\
\includegraphics[width=0.8\textwidth]{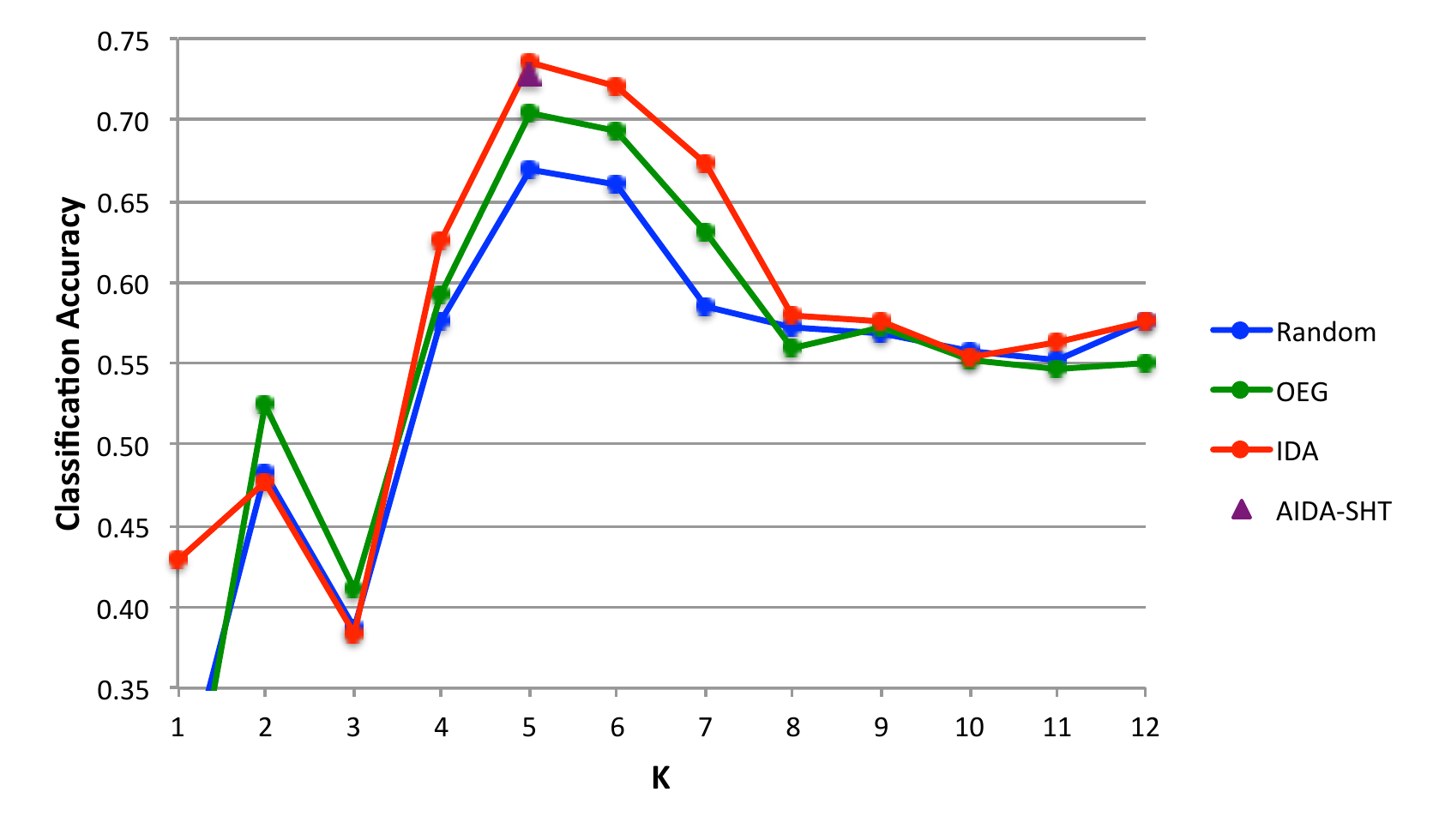}\\
(b)
\end{center}
\caption{Classification accuracy for the satellite dataset (CS to 6 samples) using a) dictionary learning, b) best learned dictionary.}
\label{Figure_6}
\end{figure}

In this section we test on the Statlog (Landsat Satellite) dataset, consisting of six classes with 36 numerical attributes and 6435 instances \cite{UCIMLR}. For this dataset, we used dictionary learning  and data-adaptation. As Figure \ref{Figure_6}a shows, only IDA achieves good classification accuracy when training the dictionaries (GMM with 6 classes). Hence, in order to remove the effect of dictionary adaptation, we used the best learned dictionary for this dataset (obtained using IDA) for all the remaining sensing matrices. Figure \ref{Figure_6}b shows the classification accuracy using this GMM, IDA and AIDA-SHT achieve the best classification performance. AIDA-SHT, automatically stops at five measurements on average.   

These classification results correspond to the classification of each CS signal independently of the other CS signals, and within the proposed two-step framework. Therefore, it cannot be directly compared with classifiers that use all the CS data at the same time.\footnote{If we take, however, all the CS signals and use a quadratic Bayesian classifier, we obtain a classification accuracy of 82\% that is similar to the results reported in \cite{Nenadic}.} These results show that classification accuracy closely follows the results obtained using synthetic data, but now, for real data. In addition, and if we are only interested in sensing a given class, we can stop sensing that signal if we determine in the first step that the class is of no interest, reducing thus the average number of measurements. Indeed, the total number of measurements for single step (adaptive or not) sensing, with $S$ signals and $M$ measurements, is $SM$, but, if we are only interested in a given class $\gamma$, the average number of measurements reduces to $S(K (1 - p(\gamma) ) + M p(\gamma))$, where $p(\gamma)$ is the probability of class $\gamma$.

Finally, the proposed  AIDA or AIDA-SHT could be used for classification purposes only (single step), where the final classification is achieved using an offline classifier that uses all the CS data, such as support vector machines, neural networks, or Bayesian classifiers.

\section{Conclusion}
\label{sec_conclusion}
We have developed a novel two-step SCS framework tailored to GMMs, from where several SCS protocols can be chosen (Table \ref{Table1}). The best sensing protocol, in terms of accuracy of classification in the first step, reconstruction in the second step, and automatic selection of number of samples, is the AIDA-SHT in the first step and the optimal (MI) adaptive sensing in the second step. A good alternative to AIDA-SHT in the first step is a batch non-adaptive IDA or a batch AIDA with a predefined number of measurements, which seems more robust to noise and deviations from the GMM assumption. The two-step SCS is clearly superior to a batch single step SCS, the reconstruction accuracy is equal or better than a single step SCS and the average total number of measurements is lower, we can stop sensing a given signal if in the first step we determine that the signal is of no interest.  

The two-step SCS framework presented here also applies to a single step adaptive or batch sensing approach, for the particular case when $K=M$. More general GMMs than the single Gaussian per signal studied here, or other non-Gaussian models, can also be considered within this framework, at the cost of higher mathematical and possibly computational complexity. 

\appendices
\section{Solution to Equation (\ref{Eq14})}
\label{App1}
The general orthogonal Procrustes problem is defined as \cite{Procrustes}
\begin{equation}
\mathbf{X} = \mbox{argmin}_{\mathbf{\hat{X}}} \| \mathbf{A\hat{X}} - \mathbf{C} \| ^2_F, ~ s.t. ~ \mathbf{\hat{X} \hat{X}}^T = \mathbf{I}.
\label{Eq_A1}
\end{equation}

\noindent Let $\mathbf{M} = \mathbf{A}^T \mathbf{C}$, and $\mathbf{M} = \mathbf{U \Delta W}^T$ its eigen-decomposition. The solution to (\ref{Eq_A1}) is given by  $\mathbf{X} = \mathbf{U W}^T$. Now, noticing that $\| \mathbf{A} \|^2_F = \| \mathbf{A}^T \|^2_F$, we can rewrite (\ref{Eq14}) as
\begin{align}
\mathbf{B}^T = \mbox{argmin}_{ \mathbf{ \hat{B} } } \Big\{ \Big \| [  \mathbf{E}^T \mathbf{\hat{B}}^T - \mathbf{I}_N \Big \|^2_F \Big\}
~ s.t. ~ \mathbf{\hat{B}^T \hat{B}} = \mathbf{I}_N.
\label{Eq_A2}
\end{align}

Comparing (\ref{Eq_A2}) and (\ref{Eq_A1}) one can see that $\mathbf{M} = \mathbf{E}$, and if $\mathbf{E} = \mathbf{U \Delta W}^T$, then the solution to (\ref{Eq_A2}) is $\mathbf{B}^T = \mathbf{U W}^T$, hence, $\mathbf{B} = \mathbf{W U}^T$.

\section{Derivation of Equation (\ref{Eq18})}
\label{App2}
By Bayes' theorem, we know that
\begin{equation}
p( \mathbf{y}_k | g, \mathbf{y}_{(k-1)} ) =
\frac{ p( \mathbf{y}_k, g | \mathbf{y}_{(k-1)} ) } { p(g | \mathbf{y}_{(k-1)} ) }.
\label{Eq_B1}
\end{equation}

\noindent Hence, (\ref{Eq17}) becomes

\begin{equation}
I(\mathbf{y}_{k} ; g | \mathbf{y}_{(k-1)}) = E \Big \{ \log \frac{ p( \mathbf{y}_k | g, \mathbf{y}_{(k-1)} ) } { p( \mathbf{y}_k | \mathbf{y}_{(k-1)} )}  \Big \}. 
\label{Eq_B2}
\end{equation}

\noindent Also by Bayes,
\begin{equation}
p( \mathbf{y}_k | g, \mathbf{y}_{(k-1)} ) = \frac{p( \mathbf{y}_k, \mathbf{y}_{(k-1)} | g)}{p(\mathbf{y}_{(k-1)} | g)} 
= \frac{p( \mathbf{y}_{(k)} | g)}{p(\mathbf{y}_{(k-1)} | g)},
\label{Eq_B3}
\end{equation}

\begin{equation}
p( \mathbf{y}_k | \mathbf{y}_{(k-1)} ) = \frac{p( \mathbf{y}_k, \mathbf{y}_{(k-1)})}{p(\mathbf{y}_{(k-1)})} 
= \frac{p( \mathbf{y}_{(k)})}{p(\mathbf{y}_{(k-1)})}.
\label{Eq_B4}
\end{equation}

\noindent Replacing (\ref{Eq_B3}) and (\ref{Eq_B4}) in (\ref{Eq_B2}), and applying logarithm properties,
\begin{eqnarray}
I(\mathbf{y}_{k} ; g | \mathbf{y}_{(k-1)}) = E \{ \log~p( \mathbf{y}_{(k)} | g ) \}  -E \{ \log~p( \mathbf{y}_{(k)} ) \}  \nonumber \\
							      - \big( E\{ \log~p( \mathbf{y}_{(k-1) | g}  ) \} - E \{ \log~p( \mathbf{y}_{(k-1)} ) \} \big ). 
\label{Eq_B5}
\end{eqnarray}

\noindent Applying the definition of differential entropy, $H(f) = -E \{ \log~f \} $, we arrive to Equation (\ref{Eq18}).

\section{Derivation of Equation (\ref{Eq27})}
\label{App3}
The covariance of $\mathbf{y}_{(k)}$, given Gaussian class $g$, is given by
\begin{align}
 | \mathbf{\Sigma}_{\mathbf{y}_{(k)} | g} | &= | \mathbf{\Phi}_{(k)} \mathbf{\Sigma}_g \mathbf{\Phi}^T_{(k)}  + \sigma^2 \mathbf{I}_{kb} | 
= \Bigg | 
\begin{bmatrix} 
\mathbf{\Phi}_{(k-1)} \\
\mathbf{\Phi}_k
\end{bmatrix}			     	
\mathbf{\Sigma}_g \big [ \mathbf{\Phi}^T_{(k-1)} ~  \mathbf{\Phi}^T_k \big ]+ \sigma^2 \mathbf{I}_{kb}  
\Bigg | \nonumber \\
&= 
\begin{vmatrix} 
\mathbf{\Phi}_{(k-1)} \mathbf{\Sigma}_g \mathbf{\Phi}^T_{(k-1)} + \sigma^2 \mathbf{I}_{(k-1)b} & \mathbf{\Phi}_{(k-1)} \mathbf{\Sigma}_g \mathbf{\Phi}^T_k\\
\mathbf{\Phi}_k \mathbf{\Sigma}_g \mathbf{\Phi}^T_{(k-1)} & \mathbf{\Phi}_k \mathbf{\Sigma}_g \mathbf{\Phi}^T_k + \sigma^2 \mathbf{I}_b
\end{vmatrix} \nonumber 
=
\begin{vmatrix} 
\mathbf{\Sigma}_{\mathbf{y}_{(k-1)}  | g } & \mathbf{\Phi}_{(k-1)} \mathbf{\Sigma}_g \mathbf{\Phi}^T_k\\
\mathbf{\Phi}_k \mathbf{\Sigma}_g \mathbf{\Phi}^T_{(k-1)} & \mathbf{\Phi}_k \mathbf{\Sigma}_g \mathbf{\Phi}^T_k + \sigma^2 \mathbf{I}_b
\end{vmatrix}. \nonumber  
\end{align}

\noindent Since by hypothesis $\mathbf{x}$ is an $N$-dimensional Gaussian with positive definite covariance $\mathbf{\Sigma}_g$ ($g$ unknown of course), then the covariance matrix $\mathbf{\Phi}_{(k-1)} \mathbf{\Sigma}_g \mathbf{\Phi}^T_{(k-1)} $ is also positive definite \cite{ProbabilityBook}, hence, invertible and we can use the identity for determinants
$\begin{vmatrix}
\mathbf{A} & \mathbf{B} \\
\mathbf{C} & \mathbf{D} 
\end{vmatrix}
= |\mathbf{A}| | \mathbf{D} - \mathbf{C} \mathbf{A}^{-1} \mathbf{B} |$
\cite{MatrixCookbook, MatrixRef}.
Then,
\begin{align}
| \mathbf{\Sigma}_{\mathbf{y}_{(k)} | g} | &=  | \mathbf{\Sigma}_{\mathbf{y}_{(k-1)}  | g } | ~ | \mathbf{\Phi}_k \mathbf{\Sigma}_g \mathbf{\Phi}^T_k + \sigma^2 \mathbf{I}_b -  \mathbf{\Phi}_k \mathbf{\Sigma}_g (\mathbf{\Phi}_{(k-1)})^T \mathbf{\Sigma}_{\mathbf{y}_{(k-1)}  | g }^{-1} \mathbf{\Phi}_{(k-1)} \mathbf{\Sigma}_g \mathbf{\Phi}^T_k  | 
\label{Eq_C1} \\
&=  | \mathbf{\Sigma}_{\mathbf{y}_{(k-1)}  | g } |  ~ \Big | \mathbf{\Phi}_k \big ( \mathbf{\Sigma}_g - \mathbf{\Sigma}_g  \mathbf{\Phi}^T_{(k-1)} \mathbf{\Sigma}_{\mathbf{y}_{(k-1)}  | g }^{-1} \mathbf{\Phi}_{(k-1)} \mathbf{\Sigma}_g + \sigma^2 \mathbf{I}_b ) \mathbf{\Phi}^T_k \Big |, \nonumber
\end{align}

\noindent where in the last step we used the fact that $\mathbf{I}_b = \mathbf{\Phi}_k \mathbf{\Phi}^T_k$. 

Starting from (\ref{Eq24}), and following the same steps indicated before, it is straightforward to show that
\begin{equation}
| \mathbf{\bar{\Sigma}}_{\mathbf{y}_{(k)}} | = 
 | \mathbf{\bar{\Sigma}}_{\mathbf{y}_{(k-1)}} |  \Big | \mathbf{\Phi}_k \big ( \mathbf{\bar{\Sigma}} - \mathbf{\bar{\Sigma}}  \mathbf{\Phi}^T_{(k-1)} \mathbf{\bar{\Sigma}}_{\mathbf{y}_{(k-1)}}^{-1} \mathbf{\Phi}_{(k-1)} \mathbf{\bar{\Sigma}} + \sigma^2 \mathbf{I}_b ) \mathbf{\Phi}^T_k \Big |. 
\label{Eq_C2}
\end{equation}

\noindent Replacing (\ref{Eq_C1}) and (\ref{Eq_C2}) in (\ref{Eq25}), and defining $\mathbf{P}_g$ and $\mathbf{\bar{P}}$ as indicated on (\ref{Eq27}), we arrive at Equation (\ref{Eq27}).

\section{Derivation of Equation (\ref{Eq34})}
\label{App4}
By Bayes' theorem,
\begin{equation}
\frac{p( \mathbf{y}^\gamma_k, \mathbf{x}^\gamma | \mathbf{y}^\gamma_{(k-1)} )}{p(\mathbf{x}^\gamma | \mathbf{y}^\gamma_{(k-1)} )} 
=p( \mathbf{y}^\gamma_k | \mathbf{x}^\gamma, \mathbf{y}^\gamma_{(k-1)} )
=\frac{p( \mathbf{y}^\gamma_k, \mathbf{y}^\gamma_{(k-1)}| \mathbf{x}^\gamma)}{p(\mathbf{y}^\gamma_{(k-1)} |  \mathbf{x}^\gamma)} 
=\frac{p( \mathbf{y}^\gamma_{(k)}| \mathbf{x}^\gamma)}{p(\mathbf{y}^\gamma_{(k-1)} |  \mathbf{x}^\gamma)} . 
\label{Eq_D1}
\end{equation}

\noindent Also, by Bayes,
\begin{equation}
p( \mathbf{y}^\gamma_k | \mathbf{y}^\gamma_{(k-1)} )= \frac{p( \mathbf{y}^\gamma_k, \mathbf{y}^\gamma_{(k-1)} ) }{p(\mathbf{y}^\gamma_{(k-1)} )}
=  \frac{p(\mathbf{y}^\gamma_{(k)} )}{p(\mathbf{y}^\gamma_{(k-1)} )}.
\label{Eq_D2}
\end{equation}
 
 \noindent Replacing (\ref{Eq_D1}) and (\ref{Eq_D2}) in (\ref{Eq34}),
\begin{equation}
I(\mathbf{y}^\gamma_k; \mathbf{x}^\gamma  | \mathbf{y}^\gamma_{(k-1)}) =
E \Big \{ \log \frac{ p( \mathbf{y}^\gamma_{(k)}| \mathbf{x}^\gamma) p(\mathbf{y}^\gamma_{(k-1)} )} { p(\mathbf{y}^\gamma_{(k-1)} |  \mathbf{x}^\gamma) p(\mathbf{y}^\gamma_{(k)} )}  \Big \},
\label{Eq_D3}
\end{equation}

\noindent and applying logarithm properties and the definition of differential entropy,
\begin{equation}
I(\mathbf{y}^\gamma_k; \mathbf{x}^\gamma  | \mathbf{y}^\gamma_{(k-1)}) =
H \big( p( \mathbf{y}^\gamma_{(k)} ) \big) - H \big( p( \mathbf{y}^\gamma_{(k)} | \mathbf{x}^\gamma) \big) - \Big[ H \big( p( \mathbf{y}^\gamma_{(k-1)} ) \big) - H \big( p( \mathbf{y}^\gamma_{(k-1)} | \mathbf{x}^\gamma) \big) \Big ]. 
\label{Eq_D4}
\end{equation}

The term within brackets in (\ref{Eq_D4}) is independent of $\mathbf{\Phi}_k$, so it can be considered a constant. The probability $p( \mathbf{y}^\gamma_{(k)} ) = p( \mathbf{y}_{(k)} | g = \gamma)$ is the same as (\ref{Eq19}), replacing $g$ by $\gamma$, and its entropy is given by \cite{MI_book}
\begin{equation}
H \big ( p( \mathbf{y}^\gamma_{(k)}) \big) =  \frac{1}{2} \Big [ kb(1 + \log(2 \pi)) + \log | \mathbf{\Sigma}_{\mathbf{y}^\gamma_{(k)}}  |  \Big ], ~
\mathbf{\Sigma}_{\mathbf{y}^\gamma_{(k)}} = \mathbf{\Phi}_{(k)} \mathbf{\Sigma}_\gamma \mathbf{\Phi}^T_{(k)}.
\label{Eq_D5}  
\end{equation}
  
\noindent On the other hand, 
\begin{equation}
 p( \mathbf{y}^\gamma_{(k)} | \mathbf{x}^\gamma) = \frac{1}{ (2 \pi \sigma^2)^{kb/2}} exp \Big ( -\frac{1}{2 \sigma^2} (\mathbf{y}^\gamma_{(k)} - \mathbf{\Phi}_{(k)} \mathbf{x}^\gamma )^T (\mathbf{y}^\gamma_{(k)} - \mathbf{\Phi}_{(k)} \mathbf{x}^\gamma ) \Big ),
\label{Eq_D6}  
\end{equation}

\noindent which has entropy $H(p( \mathbf{y}^\gamma_{(k)} | \mathbf{x}^\gamma)) = \frac{kb}{2} \log (2 \pi \sigma^2)$, and is independent of $\mathbf{\Phi}_k$. Hence, Equation (\ref{Eq33}) can be rewritten as,
\begin{equation}
I(\mathbf{y}^\gamma_k; \mathbf{x}^\gamma  | \mathbf{y}^\gamma_{(k-1)})
=  \frac{1}{2} \log | \mathbf{\Sigma}_{\mathbf{y}^\gamma_{(k)}} | + F(\mathbf{y}^\gamma_{(k-1)}),
\label{Eq_D7}
\end{equation}

\noindent where $F(\mathbf{y}^\gamma_{(k-1)})$ accounts for all terms dependent on $\mathbf{y}^\gamma_{(k-1)}$ plus some constants. Hence, in order to solve (\ref{Eq32}), we need to maximize $\log | \mathbf{\Sigma}_{\mathbf{y}^\gamma_{(k)}} | $. Since the logarithm is a monotonic function, it is enough to maximize $| \mathbf{\Sigma}_{\mathbf{y}^\gamma_{(k)}} | $. Now, following the same steps indicated in (\ref{Eq_C1}), we arrive at
\begin{equation}
| \mathbf{\Sigma}_{\mathbf{y}^\gamma_{(k)}} | =
| \mathbf{\Sigma}_{\mathbf{y}^\gamma_{(k-1)}} |  ~ \Big | \mathbf{\Phi}_k \big ( \mathbf{\Sigma}_\gamma - \mathbf{\Sigma}_\gamma  \mathbf{\Phi}^T_{(k-1)} \mathbf{\Sigma}_{\mathbf{y}^\gamma_{(k-1)}}^{-1} \mathbf{\Phi}_{(k-1)} \mathbf{\Sigma}_\gamma + \sigma^2 \mathbf{I}_b ) \mathbf{\Phi}^T_k \Big |. 
\label{Eq_D8} 
\end{equation}

\noindent Let $ \mathbf{P}_\gamma = \mathbf{\Sigma}_\gamma - \mathbf{\Sigma}_\gamma \mathbf{\Phi}^T_{(k-1)} \mathbf{\Sigma}^{-1}_{\mathbf{y}^\gamma_{(k-1)}} \mathbf{\Phi}_{(k-1)} \mathbf{\Sigma}_\gamma + \sigma^2 \mathbf{I}_b$, and since $| \mathbf{\Sigma}_{\mathbf{y}^\gamma_{(k-1)}} |$ is independent of  
$\mathbf{\Phi}_k$, the solution to (\ref{Eq32}), reduces to,
\begin{equation}
\mathbf{\Phi}_k = \mbox{argmax}_{\mathbf{\hat{\Phi}}_k} \Big | \mathbf{\hat{\Phi}}_k \mathbf{P}_\gamma \mathbf{\hat{\Phi}}^T_k \Big |, 
~s.t.~ \mathbf{\hat{\Phi}}_k \mathbf{\hat{\Phi}}^T_k = \mathbf{I}.
\label{Eq_D9}
\end{equation}

This corresponds to a high dimensional extension of the Rayleigh-Ritz Theorem, whose solution is given by Equation (\ref{Eq34}) (see proof of this theorem in \cite{MatrixRef}). 

\section{Synthetic Multidimensional Gaussian Distributions}
\label{App5}
Let $\mathbf{R}$ be an $N$$\times$$N$ Gaussian random matrix and $\mathbf{R}=\mathbf{U \Delta V}^T$ its corresponding singular value decomposition. We define synthetic covariance matrices as
$\mathbf{\Sigma} = \mathbf{U \Lambda U}^T, ~ \mathbf{\Lambda}_{ii} =  r 10^\beta i^{-\omega}$, 
where $r \in (0 ~ 1], \beta \in [4 8], \omega \in \{3, 4\}$. These ranges where chosen empirically, in order to obtain Bhattacharyya distances (BDs) in the range $[30 ~ + \infty]$, where the BD is given by
$BD = \frac{1}{2} \ln (\frac{|\Sigma_0|}{\sqrt{|\Sigma_0||\Sigma_1|}})$.

More specifically, we obtained around 100 pairs of covariances matrices with BDs for each one of the following ranges: [30 46), [46  62), [62 78), [78 94), [94 110], (110 126], (126 142], and $(142 ~ +\infty)$, for a total of 800 pairs of covariance matrices. Bhattacharyya distances among patches from natural images are in the [30 61] range, which justifies the chosen ranges. Larger Bhattacharyya distances than those considered here are quite difficult to obtain at random and often lead to numerical instability.

\noindent
{\bf Acknowledgments:}
Work supported by ONR, NSF, DARPA, NGA, ARO, and NSSEFF. We thank very constructive comments and discussions with Prof. Robert Calderbank, Prof. David Brady, and Prof. Stanley Osher.
\ifCLASSOPTIONcaptionsoff
  \newpage
\fi

\bibliographystyle{IEEEtran}
\bibliography{Bibliography}

\newpage

\end{document}